\newif\ifneurips \neuripsfalse
\newif\ificml \icmlfalse
\newif\ificlr \iclrfalse
\newif\ifcolm \colmfalse
\newif\ifacl \aclfalse

\newif\iftonyedit \tonyedittrue

\documentclass[11pt]{article}
\usepackage[preprint]{sty/acl/template} 
\acltrue

\usepackage{silence}
\PassOptionsToPackage{usenames,dvipsnames}{xcolor}
\PassOptionsToPackage{breaklinks,colorlinks}{hyperref}

\usepackage{amsmath,amsopn}

\ifcolm
\usepackage{amssymb,mathtools,amsthm,lastpage,relsize}
\else
\usepackage{amssymb,mathtools,amsthm,fancyhdr,lastpage,times}
\fi

\usepackage{algorithm}
\usepackage{algorithmic}
\usepackage{latexsym}
\usepackage{subfigure}
\usepackage{endnotes,microtype,xspace,graphicx,fancyvrb,multirow}
\usepackage{booktabs}
\usepackage{array,underscore,relsize}
\usepackage[utf8]{inputenc}
\usepackage[T1]{fontenc}
\usepackage{enumitem}
\usepackage{pifont}
\ifacl
  \usepackage[skip=5pt]{caption}
\else
  \usepackage[labelfont=bf,font=small,skip=5pt]{caption}
\fi
\usepackage{lipsum}

\usepackage[textsize=tiny]{todonotes}

\usepackage{url}

\ifneurips
\PassOptionsToPackage{numbers,sort&compress}{natbib}
\else\ificml
\PassOptionsToPackage{numbers,sort&compress}{natbib}
\else\ificlr
\PassOptionsToPackage{numbers,sort&compress}{natbib}
\else
\PassOptionsToPackage{square,comma,numbers,sort&compress}{natbib}
\fi\fi\fi

\usepackage{fp}
\usepackage{siunitx}

\usepackage{minted}

\usepackage{balance}

\usepackage{lineno}
\usepackage{hyperref}
\definecolor{mydarkblue}{rgb}{0,0.08,0.45}
\hypersetup{
  citecolor=mydarkblue,
  linkcolor=mydarkblue,
  breaklinks=true,
  urlcolor=mydarkblue
}
\usepackage[capitalize,noabbrev]{cleveref}

\usepackage{tabularx}
\usepackage{multicol}
\usepackage{makecell}
\usepackage{adjustbox}
\usepackage[normalem]{ulem}
\usepackage{tcolorbox}
\usepackage{etoc}

\microtypecontext{spacing=nonfrench}

\newcommand{\ie}{\textit{i.e.}}
\newcommand{\eg}{\textit{e.g.}}

\newcommand{\ft}{\texttt{[TAKE\_FLOOR]}}
\newcommand{\bc}{\texttt{[BACKCHANNEL]}}
\newcommand{\na}{\texttt{[LISTEN]}}

\newcommand{\framework}{\textsc{DuplexGen}\xspace} 

\newcommand{\sd}[1]{%
    \hspace{0.08em}%
    \textcolor{black!55}{%
        \raisebox{0.05ex}{%
            \fontsize{6.5}{7.2}\selectfont
            $\pm\,#1$%
        }%
    }%
}
\newcommand{\sdrow}[1]{%
    \textcolor{black!55}{\scriptsize $\pm\,#1$}%
}

\usepackage{glossaries} 
\usepackage{hyperref}
\glsdisablehyper
\usepackage{graphicx}
\usepackage[dvipsnames]{xcolor}
\usepackage{float}
\usepackage{xspace}
\usepackage{algorithm}
\usepackage{array}
\usepackage{eurosym}
\usepackage{wrapfig}
\usepackage{twemojis}

\definecolor{illiniorange}{HTML}{FF5F05}

\newacronym{llm}{LLM}{large language model}
\newacronym{sds}{SDS}{spoken dialogue system}
\newacronym{sd}{SD}{Spoken Dialogue}
\newacronym{tts}{TTS}{text-to-speech}

\usepackage{microtype}
\usepackage{graphicx}
\usepackage{subfigure}
\usepackage{booktabs}
\usepackage{amssymb}

\usepackage{tabularx}
\usepackage{multicol}
\usepackage{multirow}
\usepackage{minted}
\setminted{
  breaklines=true,
  breaksymbol={},
  breaksymbolleft={},
  breaksymbolright={},
  linenos=false,
  bgcolor=gray!10,
  frame=leftline,
  framerule=0.5pt,
  framesep=3mm,
}

\usepackage[
  commandnameprefix=always,
  todonotes={colorinlistoftodos,prependcaption,textsize=tiny,textwidth=0.83in}
]{changes}

\definechangesauthor[color=Green,name=Kangwook]{KW}
\definechangesauthor[color=Blue,name=Takyoung]{TK}
\definechangesauthor[color=Orange,name=Dilek]{DHT}
\definechangesauthor[color=Purple,name=Julia]{JH}

\definecolor{illiniorange}{HTML}{FF5F05}
\NewDocumentCommand{\tbu}
{ mO{} }{\textcolor{illiniorange}{\textsuperscript{\textit{TBU}}\textsf{\small{#1}}}}

\NewDocumentCommand{\tk}
{ mO{} }{\textcolor{blue}{\textsuperscript{\textit{TBU}}\textsf{\small{#1}}}}
\NewDocumentCommand{\kw}
{ mO{} }{\textcolor{orange}{\textsuperscript{\textit{KW}}\textsf{\small{#1}}}}
\NewDocumentCommand{\dht}
{ mO{} }{\textcolor{Green}{\textsuperscript{\textit{DHT}}\textsf{\small{#1}}}}
\NewDocumentCommand{\jh}
{ mO{} }{\textcolor{Purple}{\textsuperscript{\textit{JH}}\textsf{\small{#1}}}}

\usepackage{amsmath}
\usepackage{amssymb}
\usepackage{mathtools}
\usepackage{amsthm}

\usepackage[capitalize,noabbrev]{cleveref}

\theoremstyle{plain}

\theoremstyle{definition}

\theoremstyle{remark}

\fvset{fontsize=\scriptsize,xleftmargin=8pt,numbers=left,numbersep=5pt}

\input{code/fmt}

\def\Snospace~{\S{}}

\newcommand{\x}{$\times$\xspace}

\newif\ifdraft\drafttrue
\newif\ifnotes\notestrue
\ifdraft\else\notesfalse\fi

\ifx\pdfgentounicode\undefined
\else
  \input{glyphtounicode}
\fi

\newcolumntype{R}[1]{>{\raggedleft\let\newline\\\arraybackslash\hspace{0pt}}p{#1}}

\newcommand{\includepdf}[1]{
  \includegraphics[width=\columnwidth]{#1}
}

\newcommand{\squishlist}{
\begin{itemize}[noitemsep,nolistsep]
  \setlength{\itemsep}{-0pt}
}
\newcommand{\squishend}{
  \end{itemize}
}

\usepackage{tikz}

\usepackage{xstring}
\newcommand{\PP}[1]{
\vspace{2px}
\noindent{\bf \IfEndWith{#1}{.}{#1}{#1.}}
}

\newcommand{\boxbeg}{
\vspace{2px}
\noindent\begin{tabular}{|l|}\hline
\begin{minipage}{3.2in}
\vspace{2px}
\noindent
}

\newcommand{\boxend}{
\vspace{2px}
\end{minipage}\\ \hline
\end{tabular}
\vspace{-10pt}
}

\begin{document}

\addtocontents{toc}{\protect\setcounter{tocdepth}{-10}}

\ificml
\else
	\title{{\framework: Adaptive Synthesis of Human--AI Turn-Taking Dialogues}}

	\ifdefined\DRAFT
		\pagestyle{fancyplain}
		\lhead{Rev.~\therev}
		\rhead{\thedate}
		\cfoot{\thepage\ of \pageref{LastPage}}
	\fi


	\author{
		Takyoung Kim$^{1}$\thanks{Equal contribution.}\quad
		Kang-wook Kim$^{2,4}$\footnotemark[1]\quad
		Sang Hoon Woo$^{2,5}$ \\
		\textbf{Julia Hirschberg}$^{3}$\quad
		\textbf{Gunhee Kim}$^{2}$\quad
		\textbf{Dilek Hakkani-Tür}$^{1}$
		\\ \\[-0.5em]
		{$^1$University of Illinois Urbana-Champaign} \,
		{$^2$Seoul National University} \,
		{$^3$Columbia University} \\
		{$^4$University of California, Berkeley} \;
		{$^5$Georgia Institute of Technology} \\
		\url{tk30@illinois.edu} \qquad
		\url{kangwook@berkeley.edu}
        \\ \\[-0.5em]
        \raisebox{-0.1em}{\twemoji[height=1em]{house}}
 Project Website: \href{https://duplexgen.github.io}{\textcolor{illiniorange}{\texttt{duplexgen.github.io}}}
	}





	\date{}
	\maketitle
	\renewcommand{\thefootnote}{}
	\renewcommand{\thefootnote}{\arabic{footnote}}

	\sloppy
\fi

\begin{abstract}
Turn-taking is a central component of full-duplex interaction. Which turn-taking behaviors are appropriate varies with the scenario, yet current models apply a single norm regardless of context. This limitation originates in their training data: human--human speech corpora capture natural timing phenomena but provide little role grounding or scenario-specific norms, while heuristic or prompted synthesis methods inject turn-taking behaviors without basing them on human preferences. We introduce \textbf{\framework}, a framework for generating dialogues with scenario-adaptive turn-taking by calibrating \acrshort{llm} predictions against a small set of slot-level human preference annotations.
In six cooperative and competitive tasks, human turn-taking preferences differ systematically, and \framework aligns substantially more closely with those preferences than uncalibrated prompting or training solely on generic human--human data; a full-duplex model trained on \framework-generated data exhibits distinctive, human-preferred turn-taking behaviors. These results show that human calibration, not corpus scale or prompt design alone, is what allows turn-taking synthesis to be scenario-specific.
\end{abstract}

\glsreset{llm}

\section{Introduction}

Full-duplex language models~\citep{defossez2024moshi, yu2025salmonnomni} support concurrent listening and speaking, offering an alternative to traditional turn-based systems. A central component of this setting is \emph{turn-taking}, the moment-to-moment coordination of who holds the conversational floor~\citep{SACKS19787}. Addressing this challenge requires the assistant to make continual decisions throughout the interaction: whether to take, hold, or yield the floor, and whether to acknowledge the user without claiming it~\citep{wang2024a, zhang-etal-2024-beyond, veluri-etal-2024-beyond}.

\begin{figure}[t!]
    \centering
    \includegraphics[width=\linewidth]{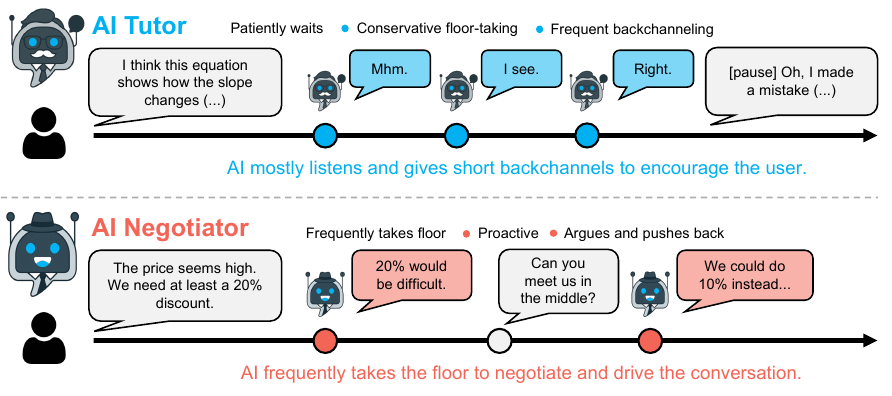}
    \captionsetup{font=small,labelfont={small}}
    \caption{Scenario-specific turn-taking behaviors.}
    \label{fig:teaser}
\end{figure}

\begin{figure*}[t!]
    \centering
    \includegraphics[width=\textwidth]{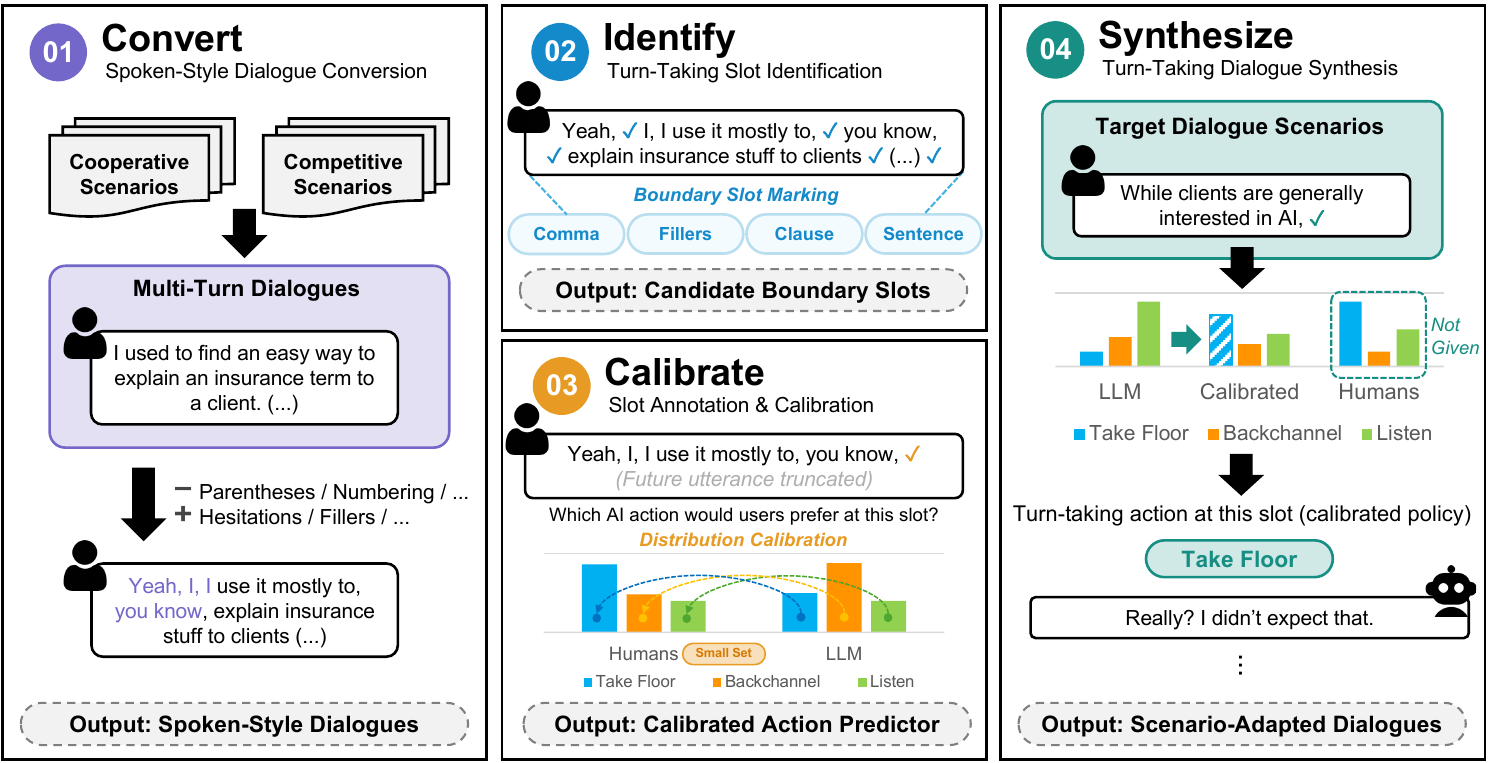}
    \captionsetup{font=small,labelfont={small}}
    \caption{Overview of the \framework framework. (1) \textbf{Spoken-style dialogue conversion:} converting text-based dialogue to spoken-style transcript format. (2) \textbf{Turn-taking slot identification:} identifying potential turn-taking slots within utterances. \;\; (3) \textbf{Slot annotation and calibration:} using a small set of human and \acrshort{llm} annotations to calibrate turn-taking predictions. \;\;\; (4) \textbf{Turn-taking dialogue synthesis:} applying the calibrated model when generating turn-taking behaviors on target scenarios.}
    \label{fig:overview}
\end{figure*}

No single turn-taking strategy is universally appropriate. Conversation analysis has long established that turn-taking behaviors depend on the social context and role structure of an interaction~\citep{SACKS19787, SKANTZE2021101178}. As \autoref{fig:teaser} illustrates, a tutor listens patiently and offers brief encouragements while a negotiator interjects more readily. Yet current full-duplex language models do not reflect this \emph{scenario-sensitivity}.

This limitation stems from the data used to train such models. Large-scale corpora of recorded human conversations, such as Switchboard~\citep{10.5555/1895550.1895693}, Fisher~\citep{cieri-etal-2004-fisher}, and Candor~\citep{doi:10.1126/sciadv.adf3197}, were built to capture natural turn-taking at scale, \textbf{at the expense of role and scenario structure}: they pair unfamiliar speakers on assigned topics, without defined task roles. An alternative is to synthesize dialogue data with explicit turn-taking actions, either through heuristic rules~\citep{chen21n_interspeech, Ma_Song_Du_Cong_Chen_Wang_Wang_Chen_2025, lin-etal-2025-predicting} or prompted \glspl{llm}~\citep{wang2024a, zhang-etal-2024-beyond, lee-etal-2025-behavior, chen-etal-2025-interactspeech}. However, the resulting turn-taking behavior is \textbf{not validated against what humans actually do in the target scenario}; \gls{llm}-based methods, in particular, inherit a turn-taking prior from pretraining~\citep{liao2025wordsmultimodalllmknows} that has not been calibrated to any specific interactional setting. Consequently, neither human-recorded nor synthetic datasets directly capture scenario-sensitivity.

We close this gap with \textbf{\framework}, a synthesis framework for scenario-adaptive turn-taking dialogue. 
As illustrated in \autoref{fig:overview}, \framework converts a text dialogue into a spoken-style dialogue, identifies candidate slots at which a listener would plausibly act, and at each slot predicts an action distribution using an \gls{llm} calibrated against a small set of slot-level human annotations per scenario. These annotations are collected as distributions over \na{}, \bc{}, and \ft{} on streaming partial utterances, rather than hard labels on complete continuations.

We instantiate \framework on six tasks spanning both cooperative and competitive interaction settings, including tutoring, planning, interview, negotiation, persuasion, and social chat. These tasks cover different goal structures and relational dynamics, allowing us to study how preferred timing changes across scenarios instead of collapsing all turn-taking into a single conversational norm.

Our experiments support a clear story. First, human turn-taking preferences differ systematically across scenarios, confirming that scenario-adaptive modeling is necessary. Second, models calibrated with a small amount of slot-level human supervision align with human judgments substantially better than prompt-only generation and models trained only on generic human--human conversation data. 
We also show that a full-duplex language model trained on a \framework-generated dataset can show distinctive turn-taking behaviors preferred by humans.
Together, these findings identify human calibration as the missing ingredient for scenario-specific turn-taking synthesis.

Our contributions are as follows.

\begin{itemize}
    \item We introduce \framework, a synthesis framework that aligns \gls{llm}-based turn-taking with scenario-specific human behavior using only a small per-scenario annotation budget.

    \item We apply \framework to six tasks spanning cooperative (tutoring, planning, interview) and competitive (negotiation, persuasion, social chat) interaction, and show that (1) a small-scale calibration enables human-aligned turn-taking dialogue synthesis, and (2) a full-duplex model trained on the resulting corpus exhibits scenario-specific turn-taking preferred by humans.
\end{itemize}

\section{Related Work}

\subsection{Turn-Taking as Scenario-Dependent Social Behavior}
\label{sec:turntaking}

Turn-taking is a foundational organizing principle of conversation. Classic conversation analysis describes interaction in terms of turn-constructional units and transition-relevance places~\citep{SACKS19787}, where listeners anticipate possible completion points and coordinate entry into the floor. Later psycholinguistic work links response timing to incremental comprehension and production~\citep{10.3389/fpsyg.2015.00731}, arguing that listeners begin preparing their responses before the current speaker has finished. Computational research has modeled turn-taking with acoustic cues, gaze, gesture, and incremental language signals~\citep{Ford_Thompson_1996, GRAVANO2011601,ekstedt-skantze-2020-turngpt}, often with the goal of minimizing latency or predicting turn boundaries.

This line of work explains \emph{how} conversational timing is coordinated, but it does not by itself resolve a central question for spoken AI assistants: \textbf{\emph{what kind of listener action is socially appropriate in a given scenario?}} The same partial user utterance may invite silence, backchanneling, or floor-taking depending on whether the assistant is acting as a tutor, planner, interviewer, or negotiation partner. \framework builds on the turn-taking literature, but focuses on the problem of modeling \emph{scenario-dependent turn-taking preferences} rather than generic endpoint prediction alone.

\subsection{Turn-Taking Dialogue Resources}

Existing data resources occupy two incomplete extremes. First, generic human--human corpora such as Switchboard~\citep{10.5555/1895550.1895693}, Fisher~\citep{cieri-etal-2004-fisher}, and Candor~\citep{doi:10.1126/sciadv.adf3197} contain natural speech and naturally occurring timing phenomena, but they are not designed around defined task roles, controlled interaction scenarios, or slot-level supervision over listener actions. They are therefore useful sources of conversational realism, but weak supervision for learning scenario-specific turn-taking norms.

Second, recent spoken-style dialogue synthesis pipelines, including Duplex-UltraChat~\citep{zhang-etal-2024-beyond}, Behavior-SD~\citep{lee-etal-2025-behavior}, and InteractSpeech~\citep{chen-etal-2025-interactspeech}, introduce turn-taking through prompting, heuristic rules, or TTS-based conversion. These approaches improve controllability and make it easier to generate large amounts of spoken-style dialogue, but they typically do not ground inserted turn-taking behaviors in human preference annotations for the target scenario. As a result, they can produce scenario-conditioned turn-taking without demonstrating that the generated behavior is actually aligned with what humans would want from the assistant.

\framework is designed to fill this gap. It combines scenario-controlled dialogue synthesis with a small set of slot-level human judgments and a context-aware calibration step before generation. In this sense, our contribution is \textbf{not simply another turn-taking dataset, but a human-calibrated synthesis framework} for producing scenario-appropriate turn-taking behaviors.

\subsection{Full-Duplex Models and Evaluation}
\label{sec:fullduplex}

Recent full-duplex speech language models have made simultaneous listening and speaking a practical target for spoken dialogue systems. This shift has motivated a growing body of benchmarks and evaluation, including FullDuplexBench~\citep{lin2025fullduplexbenchbenchmarkevaluatefullduplex}, FD-Bench~\citep{peng25b_interspeech}, Talking Turns~\citep{arora2025talking}, and FLEXI~\citep{ge2025flexibenchmarkingfullduplexhumanllm}, which measure latency, gap and overlap behavior, barge-in quality, or context-sensitive response behavior in real-time interaction.

\begin{table*}[t!]
    \small
    \captionsetup{font=small,labelfont={small}}
    \caption{Cooperative and competitive conversational tasks with reference text datasets. See \autoref{tab:statistics_data} for dataset statistics.}
    \centering 
    \renewcommand{\arraystretch}{0.9}
    \setlength{\tabcolsep}{8pt}
    \begin{tabularx}{\textwidth}{*{4}{>{\centering\arraybackslash}X}}
    \toprule
    \multicolumn{2}{c}{\textbf{Cooperative Scenario}} & \multicolumn{2}{c}{\textbf{Competitive Scenario}} \\
    \cmidrule(l{0pt}r{0pt}){1-2}\cmidrule(l{0pt}r{0pt}){3-4}
    \textbf{Turn-Taking Task} & \textbf{Reference Text Dataset} & \textbf{Turn-Taking Task} & \textbf{Reference Text Dataset} \\
    \midrule[1pt]
    
    \multirow[c]{2}{=}{\makecell[c]{Socratic Teaching\\(TEA)}} &
    SocraticLM &
    \multirow[c]{2}{=}{\makecell[c]{Negotiation\\(NEG)}} &
    CraigslistBargain \\
    & \citep{liu2024socraticlm} &
    & \citep{he-etal-2018-decoupling} \\
    \midrule

    \multirow[c]{2}{=}{\makecell[c]{Mixed-Initiative Planning\\(PLN)}} &
    MultiWOZ &
    \multirow[c]{2}{=}{\makecell[c]{Persuasion\\(PER)}} &
    DailyPersuasion \\
    & \citep{budzianowski-etal-2018-multiwoz} &
    & \citep{jin2024persuading} \\
    \midrule

    \multirow[c]{2}{=}{\makecell[c]{Interview\\(INT)}} &
    Anthropic Interviewer &
    \multirow[c]{2}{=}{\makecell[c]{Social Chat\\(SOC)}} &
    SODA \\
    & \citep{handa2025interviewer} &
    & \citep{kim-etal-2023-soda} \\
    \bottomrule

    \end{tabularx}
    \label{tab:scenario}
\end{table*}

These works can evaluate spoken dialogue systems, but they do not solve the upstream data generation problem targeted in this work. Benchmark construction often relies on generic speech corpora, black-box judging, or prompted turn-taking decisions rather than explicit human validation of which action is preferred at each candidate slot. Likewise, models trained only on human--human corpora may learn general timing regularities without learning the scenario-specific norms that govern human expectations.

\framework is complementary to this line of work. Rather than proposing another benchmark, we focus on generating the kind of training data that full-duplex assistants need but current resources do not provide: dialogues whose turn-taking behavior is explicitly calibrated to human preferences in context. This positioning places \framework at the missing corner between generic speech corpora and prompted synthetic dialogue, namely, human-calibrated, scenario-adaptive turn-taking synthesis.

\section{\framework Framework}

We focus on two linguistically expressible turn-taking behaviors, \emph{floor-taking} and \emph{backchanneling}, because they represent the fundamental dichotomy of listener responses (\ie, actively claiming versus passively acknowledging the conversational floor) and are both expressible in text, making them directly actionable for \gls{llm} assistants. Grounded in conversation-analytic theory \citep{SACKS19787}, floor-taking refers to entering at or near transition-relevance places, including competitive entries like interruptions. Backchanneling (\eg, ``uh-huh'') displays attention without claiming the floor.

\subsection{Scenario Curation}

Inspired by \citet{yang-2001-visualizing}, we distinguish between two fundamental turn-taking scenarios with distinct interactional behaviors: \emph{cooperative} and \emph{competitive}. Building upon this classification, we further diversify the scenarios to encompass a broad range of interactional contexts. As shown in \autoref{tab:scenario}, we curate six heterogeneous tasks that simulate turn-taking interactions from widely-used textual datasets. For \emph{cooperative} scenarios, we include Socratic teaching (\textbf{TEA}; guided knowledge discovery), mixed-initiative planning (\textbf{PLN}; collaborative goal completion), and interviewing (\textbf{INT}; structured information elicitation). For \emph{competitive} scenarios, we include negotiation (\textbf{NEG}; resource allocation under conflict), persuasion (\textbf{PER}; attitude change attempts), and social chat (\textbf{SOC}; peer-like conversation). Together, these six tasks span diverse goal orientations (task-driven, knowledge-seeking, social), initiative structures (system-led, user-led, mixed), and relational dynamics (expert-novice, transactional, peer-like), ensuring that the resulting turn-taking patterns generalize across the breadth of scenarios an AI assistant may encounter.

\subsection{Turn-Taking Dialogue Generation}
\label{sec:dialogue_generation}

As illustrated in \autoref{fig:overview}, \framework operates in four stages: (1) converting text-based dialogue to spoken-style transcript, (2) identifying candidate turn-taking slots, (3) calibrating \gls{llm} turn-taking decisions using a small set of human annotations, and (4) applying the learned calibration function during dialogue synthesis. In this work, we focus on modeling \textbf{the underexplored AI assistant's turn-taking behaviors} given user utterances.

We formalize a multi-turn dialogue as a sequence of turns,  where each turn comprises a sequence of text tokens. An utterance $U_D$ in dialogue turn $D$ is defined as the token sequence:
{
\setlength{\abovedisplayskip}{8pt}
\setlength{\belowdisplayskip}{8pt}
\[
    U_D = [u_D^1, u_D^2, \ldots, u_D^{n_D}],
\]
}

\noindent
where $u_D^i$ is the $i$-th token and $n_D$ the number of tokens in the $D$-th turn.

\begin{table}[t]
\captionsetup{font=small,labelfont={small}}
\caption{An example of spoken-style utterance.}
\centering
\small
\renewcommand{\arraystretch}{1.0}
\setlength{\tabcolsep}{6pt}
\begin{tabularx}{\linewidth}{XX}
\toprule
\textbf{Original} & \textbf{Spoken-Style} \\ \midrule[1pt]
I use it to find an easy way to explain an insurance term to a client. I also use it for helping write correspondence. &
\emph{Yeah, I, I} use it mostly to, \emph{you know}, explain insurance stuff to clients in a way that's easier to get. And sometimes for writing emails or letters. \\
\bottomrule[1pt]
\end{tabularx}
\label{tab:spoken}
\end{table}
\begin{table}[t]
\captionsetup{font=small,labelfont={small}}
\caption{Proportion of spoken-style aspects with comparison to Switchboard~\citep{10.5555/1895550.1895693}.}
\centering
\small
\renewcommand{\arraystretch}{1.0}
\setlength{\tabcolsep}{2pt}
\begin{tabular}{llccc}
\toprule
\textbf{Dataset} & \textbf{Role} & \textbf{Word Count} & \textbf{Filler} & \textbf{Repetition} \\ \midrule[1pt]
Switchboard & Humans & 1,518,024 & 3.88\% & 1.39\% \\ \midrule
\multirow{2}{*}{\framework} & User & 127,786 & 3.50\% & 0.43\% \\
 & Assistant & 148,430 & 0.38\% & 0.01\% \\
\bottomrule[1pt]
\end{tabular}
\label{tab:swbd}
\end{table}

\subsubsection{Spoken-Style Dialogue Conversion}
\label{sec:conversion}
We prompt an \gls{llm} to convert text-based dialogues into spoken-style format (\autoref{sec:prompt_conversion}) by removing text-based elements such as parentheses~\citep{cho-etal-2024-speechworthy} and rephrasing with speech-oriented elements such as disfluencies and filler words. \autoref{tab:spoken} shows an example of spoken-style conversion. This process ensures that the generated dialogue is suitable for speech-oriented turn-taking. Hereafter, we refer to $U_D$ as the spoken-style utterance.

To assess whether the converted spoken-style dialogues resemble natural human conversations, we compare them with the Switchboard corpus~\citep{10.5555/1895550.1895693}. We focus on sequence matching of lexically identifiable features, namely fillers and repetitions. As shown in \autoref{tab:swbd}, the converted user utterances exhibit human-like speech characteristics, especially in filler usage. Although repetitions are less frequent than in natural speech, we find this acceptable because excessive disfluency could distract from our main focus on turn-taking dynamics. By contrast, the low rate of these features in AI utterances is intentional, as the AI assistant is designed to deliver information clearly.

\subsubsection{Turn-Taking Slot Identification}
\label{sec:slot}

Upon obtaining the spoken-style utterance $U_D$, we identify candidate slots for turn-taking actions by analyzing sentence structure and linguistic boundaries. For a given dialogue context $c$ and spoken-style utterance $U_D$, we employ both heuristic rules (\autoref{sec:heuristics}) and \gls{llm} (\autoref{sec:prompt_candidate_identification}) to identify natural boundaries where turn-taking could occur. These boundaries typically correspond to clause endings, sentence completions, or other linguistic patterns like hesitations and filler words. We empirically set the maximum distance between consecutive turn-taking slots to 8 tokens, which balances annotation granularity against computational cost while roughly corresponding to typical phrase-length units in conversational speech. Let $L_D = \{l_1, l_2, \ldots, l_k\} \; (k \leq n_D)$ denote the set of potential turn-taking slots identified in utterance $U_D$. For each slot $l_j$, we define the context up to that point as $x_D^{l_j} = [u_D^1, \ldots, u_D^{l_j}]$, which will be utilized in the next step.

\subsubsection{Scenario-Specific Turn-Taking Annotation Calibration}
\label{sec:calibration}

To align turn-taking decisions with human judgments about which turn-taking actions are acceptable, we employ a calibration process that learns from the agreement between human and \gls{llm} annotations. For each scenario, we collect a small set of turn-taking annotations where humans and \glspl{llm} identify appropriate turn-taking actions respectively.

\paragraph{Annotation Collection.}
For each dialogue in the calibration set (see \autoref{sec:datasets} for details), we collect two types of annotations. First, we obtain \textbf{human annotations} $A_H = \{(l_j, a_j)\}$, where each annotation specifies a turn-taking location $l_j$ within the user utterance and the corresponding action $a_j \in \{\text{\na}, \text{\bc}, \text{\ft}\}$. Given $x_D^{l_j}$, annotators are instructed to identify appropriate turn-taking slots \emph{at which the AI can take the turn}. To mitigate the retroactive annotation bias (\ie, annotating after seeing the rest of the conversation) discussed in \citet{umair-etal-2024-large}, we present text chunks incrementally, as illustrated in \autoref{sec:demo}. We recruit 248 English-speaking participants from the United States and the United Kingdom via Prolific\footnote{\url{https://www.prolific.com}} and provide the guidelines (\autoref{sec:guideline}). Because turn-taking is inherently subjective, each sample is annotated by five participants, allowing \emph{human judgments to be treated as distributions rather than single decisions.} 

Second, we collect \textbf{\gls{llm} annotations} $A_L = \{(l_j, a_j')\}$ by predicting turn-taking action probabilities for each candidate slot. To address the miscalibration issue of recent reinforcement learning-based \glspl{llm}~\citep{tian-etal-2023-just}, we use \emph{verbalized} confidence scores (\ie, explicitly generated action distributions) to represent confidence in $a_j'$~(\autoref{sec:prompt_action_annotation}).

\paragraph{Distribution Calibration.}
\label{sec:califunction}

Given paired annotations $A_H$ and $A_L$, our goal is to learn a calibration function $f_{\text{cal}}$ that maps the \gls{llm}'s raw prediction distribution to one better aligned with human judgments. Although context-agnostic post-hoc methods such as temperature scaling can correct global miscalibration with little data, they apply the same transformation across the scenario and dialogue context, limiting their ability to capture the situational nature of turn-taking. We therefore adopt a context-aware calibration via \gls{llm} fine-tuning, conditioning on the dialogue history $c$, and partial utterance $x_D^{l_j}$ to produce calibrated predictions.

A key design choice is \textbf{the training target}. Rather than reducing human annotations for each slot to a majority-vote label, we use empirical annotation counts as a \emph{soft label}, defining the target distribution $P_H(a \mid x_D^{l_j}, c)$ by normalizing over the three action classes. This reflects the subjectivity of turn-taking: multiple actions may be appropriate at a given slot, and a hard label discards information about human disagreement. We therefore train $f_{\text{cal}}$ to align its predicted distribution $P_{\text{cal}}(a \mid x_D^{l_j}, c)$ with this soft target. The objective minimizes the KL divergence between $P_H$ and $P_{\text{cal}}$:

{
\setlength{\abovedisplayskip}{5pt}
\setlength{\belowdisplayskip}{10pt}
{\fontsize{7.8pt}{9pt}\selectfont
\[
    \mathcal{L} = \sum_{(c,\, U_D)\, \in\, \mathcal{D}_{\text{cal}}} \;\sum_{l_j\, \in\, L_D}
        D_{\mathrm{KL}}\!\left( P_H(\cdot \mid x_D^{l_j}, c) \;\Big\|\; P_{\text{cal}}(\cdot \mid x_D^{l_j}, c) \right),
\]
}
}

\noindent which encourages the model to match the full distribution of human preferences, preserving diversity in turn-taking judgments.

\subsubsection{Turn-Taking Dialogue Synthesis}
\label{sec:synthesis}

The calibrated turn-taking prediction model is used to synthesize turn-taking behaviors in target scenarios generated turn-by-turn (\autoref{sec:prompt_dial_synthesis}). Similar to the calibration phase, it scans candidate slots and decides whether to insert a turn-taking marker (\ft{} or \bc{}) for each slot\footnote{\na{} is not explicitly marked, as it is a default listener behavior of the AI.}. The insertion of \ft{} truncates the current turn and initiates the subsequent turn, whereas \bc{} adds a brief acknowledgment\footnote{The specific backchannel expression is inserted via prompting (\autoref{sec:prompt_backchannel_insertion}).} while allowing the turn to continue. The full procedure is formalized in \autoref{sec:algorithm}.

\section{Experiments}

We demonstrate experiments with diverse perspectives. Details on tasks, models, and evaluation are provided in each subsection.

\begin{figure}[t!]
    \centering
    \includegraphics[width=\linewidth]{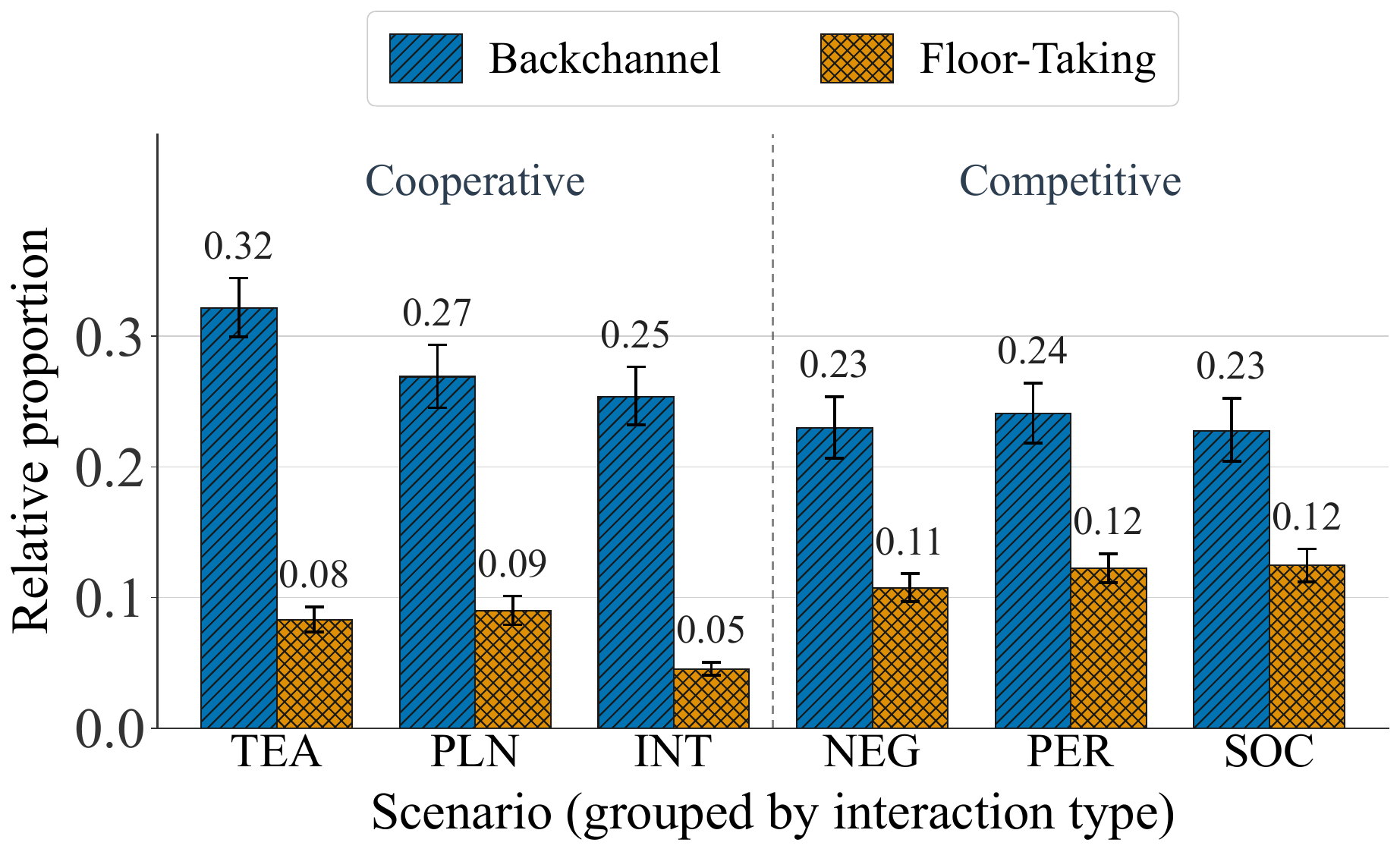}
    \captionsetup{font=small,labelfont={small}}
    \caption{Human-annotated \bc{} and \ft{} proportions per turn-taking slot, averaged across slots and grouped by cooperative and competitive interaction types. Error bars are 95\% bootstrap confidence intervals.}
    \label{fig:data_dist}
\end{figure}

\subsection{Do Turn-Taking Preferences Differ by Scenario?}
\label{sec:exp_scenario_difference}
We examine whether human turn-taking preferences vary across interaction scenarios. Using collected human annotations, we analyze the distribution of turn-taking actions across six tasks in cooperative and competitive scenarios.

\autoref{fig:data_dist} shows the relative proportions of backchanneling and floor-taking for each turn-taking slot (density plots are provided in \autoref{sec:human_annotation_distribution}). Pairwise chi-square tests with Holm correction show significant differences across nearly all scenario pairs ($p < 0.05$), except within competitive scenarios, suggesting that competitive scenarios share a common turn-taking norm.

Notably, backchanneling is more frequent than average in cooperative settings. 
Floor-taking is more common in competitive scenarios and less so in cooperative ones, as confirmed by pairwise comparisons. 
Overall, these results suggest that \textbf{cooperative and competitive interactions follow distinct turn-taking norms}, while tasks within the same scenario type may share similar patterns.

\subsection{Can Calibrated \texorpdfstring{\glspl{llm}}{LLMs} Predict Human-Like Turn-Taking?}
\label{sec:exp_turn_prediction}

We evaluate the effectiveness of the calibrated \gls{llm} in predicting human-like turn-taking behaviors.



\begin{table}[t]
\captionsetup{font=small,labelfont={small}}
\caption{Dataset statistics in \autoref{sec:exp_turn_prediction}.}
\centering \small
\renewcommand{\arraystretch}{0.5}
\setlength{\tabcolsep}{1pt}
\begin{tabularx}{\linewidth}{
    >{\raggedright\arraybackslash}X
    >{\raggedright\arraybackslash}X
}
\toprule
\multicolumn{2}{c}{\textbf{\framework-Generated Dataset Statistics}} \\ 
\midrule[1pt]

\# Dialogues (Calibration) 
& 120 (20 per task) \\ 
\midrule

\# Utterances (Calibration) 
& 240 (2 per dial.; 40 per task) \\ 
\midrule

\# Dialogues (Evaluation) 
& 300 (50 per task) \\ 
\midrule

\# Utterances (Evaluation) 
& 600 (2 per dial.; 100 per task) \\ 
\midrule

Avg. words per utterance 
& 24.57 \\ 
\midrule

\makecell[l]{Avg. turn-taking slots\\[-0.1em] per utterance} 
& 5.57 \\

\bottomrule[1pt]
\end{tabularx}
\label{tab:statistics_data}
\end{table}
\begin{table}[t]
\captionsetup{font=small,labelfont={small}}
\caption{The number of turn-taking slots for calibration and evaluation sets across six turn-taking tasks in \autoref{sec:exp_turn_prediction}.}
\centering \small
\renewcommand{\arraystretch}{1.0}
\setlength{\tabcolsep}{5.3pt}
\begin{tabularx}{\linewidth}{lcccccc}
\toprule
& \multicolumn{3}{c}{\textbf{Cooperative}} & \multicolumn{3}{c}{\textbf{Competitive}} \\
\cmidrule(lr){2-4} \cmidrule(lr){5-7}
\textbf{Split} & TEA & PLN & INT & NEG & PER & SOC \\
\midrule[1pt]
Calibration & 160 & 148 & 415 & 242 & 232 & 198 \\\midrule
Evaluation & 383 & 341 & 1016 & 439 & 577 & 497 \\
\bottomrule
\end{tabularx}
\label{tab:statistics_slot}
\end{table}

\begin{table*}[t]
    \captionsetup{font=small,labelfont={small}}
    \caption{Mean and standard error of the mean (SEM) of KL divergence on the evaluation split across six turn-taking scenarios (lower is better). Standard deviations are shown in gray. Methods are ordered from the simplest baseline to our full method.}
    \label{tab:summary_results}

    \centering
    \small
    \renewcommand{\arraystretch}{1.1}
    \setlength{\tabcolsep}{2.2pt}

    \begin{tabularx}{\linewidth}{lllllllll}
        \toprule
        & &
        \multicolumn{3}{c}{\textbf{Cooperative}} &
        \multicolumn{3}{c}{\textbf{Competitive}} & \\
        \cmidrule(lr){3-5}
        \cmidrule(lr){6-8}

        \textbf{Size}
        & \textbf{Method}
        & TEA
        & PLN
        & INT
        & NEG
        & PER
        & SOC
        & \textbf{AVG ($\downarrow$)} \\
        \midrule[1pt]

        \multirow{4}{*}{1.7B}
        & \textsc{Prompt-Only}
        & 9.492\sd{0.412}
        & 7.231\sd{0.361}
        & 4.241\sd{0.196}
        & 6.336\sd{0.346}
        & 4.550\sd{0.290}
        & 6.616\sd{0.352}
        & 6.411\sd{0.127} \\

        & \textsc{SWBD-Only}
        & 1.331\sd{0.061}
        & 1.210\sd{0.061}
        & 0.800\sd{0.030}
        & 1.420\sd{0.067}
        & 1.545\sd{0.067}
        & 1.254\sd{0.061}
        & 1.260\sd{0.022} \\

        & \framework-\textsc{Only}
        & \textbf{0.471}\sd{0.025}
        & \textbf{0.413}\sd{0.022}
        & 0.353\sd{0.012}
        & 0.505\sd{0.024}
        & \textbf{0.490}\sd{0.020}
        & 0.558\sd{0.023}
        & \textbf{0.465}\sd{0.008} \\

        & SWBD+\framework
        & 0.533\sd{0.030}
        & 0.492\sd{0.032}
        & \textbf{0.271}\sd{0.009}
        & \textbf{0.492}\sd{0.026}
        & 0.512\sd{0.024}
        & \textbf{0.494}\sd{0.025}
        & 0.466\sd{0.009} \\

        \midrule

        \multirow{4}{*}{4B}
        & \textsc{Prompt-Only}
        & 0.370\sd{0.015}
        & 0.608\sd{0.111}
        & 0.363\sd{0.009}
        & 0.475\sd{0.040}
        & 0.441\sd{0.015}
        & 0.492\sd{0.046}
        & 0.458\sd{0.015} \\

        & \textsc{SWBD-Only}
        & 0.762\sd{0.026}
        & 0.729\sd{0.026}
        & 0.647\sd{0.013}
        & 0.804\sd{0.027}
        & 0.827\sd{0.024}
        & 0.746\sd{0.022}
        & 0.752\sd{0.009} \\

        & \framework-\textsc{Only}
        & 0.558\sd{0.033}
        & 0.673\sd{0.041}
        & 0.496\sd{0.019}
        & 0.603\sd{0.038}
        & 0.587\sd{0.031}
        & 0.779\sd{0.032}
        & 0.616\sd{0.012} \\

        & SWBD+\framework
        & \textbf{0.299}\sd{0.017}
        & \textbf{0.339}\sd{0.024}
        & \textbf{0.255}\sd{0.008}
        & \textbf{0.346}\sd{0.018}
        & \textbf{0.386}\sd{0.017}
        & \textbf{0.389}\sd{0.017}
        & \textbf{0.335}\sd{0.006} \\

        \midrule

        \multirow{4}{*}{14B}
        & \textsc{Prompt-Only}
        & 0.700\sd{0.022}
        & 0.644\sd{0.028}
        & 0.508\sd{0.013}
        & 0.700\sd{0.024}
        & 0.755\sd{0.023}
        & 0.781\sd{0.025}
        & 0.681\sd{0.009} \\

        & \textsc{SWBD-Only}
        & 1.715\sd{0.075}
        & 1.380\sd{0.068}
        & 0.955\sd{0.036}
        & 1.568\sd{0.075}
        & 1.727\sd{0.072}
        & 1.485\sd{0.071}
        & 1.472\sd{0.026} \\

        & \framework-\textsc{Only}
        & 1.234\sd{0.068}
        & 0.920\sd{0.071}
        & 0.696\sd{0.027}
        & 1.127\sd{0.069}
        & 1.067\sd{0.055}
        & 0.922\sd{0.047}
        & 0.994\sd{0.021} \\

        & SWBD+\framework
        & \textbf{0.358}\sd{0.021}
        & \textbf{0.311}\sd{0.021}
        & \textbf{0.264}\sd{0.009}
        & \textbf{0.407}\sd{0.025}
        & \textbf{0.438}\sd{0.021}
        & \textbf{0.383}\sd{0.018}
        & \textbf{0.360}\sd{0.007} \\

        \bottomrule
    \end{tabularx}
\end{table*}

\subsubsection{Datasets}
\label{sec:datasets}

We sample 20 dialogues from the original training split for calibration and 50 dialogues from the original test split for evaluation for each dataset listed in \autoref{tab:scenario}.
Regarding SODA~\citep{kim-etal-2023-soda}, we further curate competitive scenarios via \gls{llm} filtering (\autoref{sec:prompt_soda_scenario_filtering}). For turn-taking annotation and prediction, we sample two utterances from each dialogue, yielding 40 calibration and 100 evaluation utterances. Because utterance lengths vary, the number of candidate turn-taking slots (\ie, the actual unit of supervised calibration) differs by scenario. 
Dataset statistics are shown in \autoref{tab:statistics_data}, and slot counts in \autoref{tab:statistics_slot}. 

\subsubsection{Models and Evaluation}

For the calibration and evaluation process of the dataset specified in \autoref{tab:statistics_data}, GPT-4.1-2025-04-14~\citep{achiam2023gpt4} is used for spoken-style dialogue conversion~(\autoref{sec:conversion}), turn-taking slot identification~(\autoref{sec:slot}), and dialogue synthesis~(\autoref{sec:synthesis}). For the calibrated turn-taking action predictor~(\autoref{sec:calibration}), we use the Qwen3 model family~\citep{yang2025qwen3}. 

Given that human turn-taking annotations were obtained for evaluation (\autoref{tab:statistics_slot}), calibration performance can be assessed directly by comparing human annotation distributions and \gls{llm} predictions. Since the annotation and prediction are conducted on a turn basis~(\autoref{sec:dialogue_generation}, \autoref{sec:demo}), the task is formulated as a turn-taking action prediction for individual user turns. We measure distributional mismatch using $D_{\mathrm{KL}}(\text{Human} \Vert \text{Model})$, comparing the human annotation distribution over turn-taking actions against the distribution predicted by the \gls{llm}.

\subsubsection{Turn-Taking Prediction Methods}

We compare the following methods using Qwen3 1.7B, 4B, and 14B models.

\begin{itemize}[leftmargin=*]
    \item \textbf{\textsc{Prompt-Only}:} A verbalized estimation method solely relying on prompts~(\autoref{sec:prompt_action_annotation}).
    \item \textbf{\textsc{SWBD-Only}:} A model fine-tuned with general human--human conversation text corpora from Switchboard that includes turn-taking markers. We add a prediction head on top of the base model to predict the turn-taking actions (see \autoref{sec:turn_taking_prediction_training} for details). SWBD contains 2.4K dialogues, which are much larger than the \framework calibration set, making it a strong baseline.
    \item \textbf{\textsc{\framework-Only}:} A model fine-tuned with a small-scale \framework calibration set. It directly learns human judgments as soft labels.
    \item \textbf{\textsc{SWBD}+\framework:} A model first fine-tuned with the text corpora of Switchboard and then fine-tuned with the \framework calibration set.
\end{itemize}

\begin{table}[t]
    \captionsetup{font=small,labelfont=small}
    \caption{
        Distribution difference and standard deviation from competitive to cooperative scenarios (4B). $\Delta$ denotes the per-action rate difference between competitive and cooperative scenarios; DG denotes \framework. A smaller difference indicates that the two scenarios are less distinguishable.
    }
        \label{tab:action_adaptation}

    \centering
    \small
    \renewcommand{\arraystretch}{1.0}
    \setlength{\tabcolsep}{2pt}

    \begin{tabular*}{\columnwidth}{
        @{\extracolsep{\fill}}
        l
        c
        r
        r
        r
        @{}
    }
        \toprule
        \textbf{Action}
        & \textbf{Metric}
        & \multicolumn{1}{c}{\textbf{Human}}
        & \multicolumn{1}{c}{\textbf{Prompt}}
        & \multicolumn{1}{c}{\textbf{SWBD+DG}} \\
        \midrule

        \multirow{2}{*}{Floor-Taking}
        & $\Delta$
        & $+$0.056
        & \textbf{$+$0.008}
        & $+$0.052 \\
        & SD
        & 0.032
        & \textbf{0.014}
        & 0.029 \\
        \midrule

        \multirow{2}{*}{Backchannel}
        & $\Delta$
        & $-$0.053
        & \textbf{$+$0.011}
        & $-$0.062 \\
        & SD
        & 0.037
        & \textbf{0.011}
        & 0.038 \\
        \bottomrule
    \end{tabular*}
\end{table}

\subsubsection{Results}
\label{sec:exp_turn_prediction_results}

\autoref{tab:summary_results} presents results against human annotations as ground truth. 
Across all methods, \textsc{SWBD+\framework} consistently achieves the lowest KL divergence\footnote{We provide an ablation study on the amount of calibration data in \autoref{sec:calibration_ablation}.} (except for 1.7B, but marginal difference\footnote{This is unsurprising, as the smaller calibration set appears sufficient for a model at that scale.}), indicating that it produces turn-taking distributions most consistent with human preferences. While the \textsc{Prompt-Only} method shows decent performance in 4B and 14B models, \textbf{this prompt-guided behavior fails to capture scenario-specific variation compared to the calibrated models}. Specifically, we compare the KL divergence difference between competitive and cooperative scenarios for \textsc{Prompt-Only} and \textsc{SWBD+\framework} using 4B model. As shown in \autoref{tab:action_adaptation}, \textsc{Prompt-Only} exhibits a smaller difference and lower standard deviation for both turn-taking actions than the proposed \textsc{SWBD+\framework} method, indicating less diversity across scenarios. In addition, we qualitatively inspect conversation examples in \autoref{sec:generated_examples} and find that \textsc{Prompt-Only} generates too many \texttt{[TAKE\_FLOOR]} slots, which might distract the user. 

Interestingly, \textsc{SWBD-Only}, despite being trained on a much larger human--human dialogue corpus, is always less aligned than \textsc{\framework-Only}. We qualitatively observe that it overestimates silence and underestimates backchanneling relative to human annotations, suggesting that \textbf{general-domain conversational data does not capture the scenario-specific turn-taking norms targeted in this work}.

\subsection{Can Synthesized Dialogues Enable Scenario-Specific Turn-Taking of Full-Duplex Model?}
\label{sec:exp_full_duplex}

We further evaluate whether \framework{} can translate calibrated turn-taking preferences into effective full-duplex dialogue behavior, beyond merely improving turn-taking prediction. Unlike slot-level prediction in \autoref{sec:exp_turn_prediction}, full-duplex interaction requires models to realize turn-taking decisions through temporally coordinated speech behavior while maintaining naturalness and adherence to the interaction scenario. We therefore conduct complementary single-turn and multi-turn analyses: the single-turn analysis examines whether models produce scenario-specific turn-taking behaviors during user speech~(\autoref{sec:duplex_tt_pattern}), while the multi-turn dialogue evaluation assesses whether these behaviors remain natural and contextually appropriate throughout an interaction~(\autoref{sec:duplex_eval}). Together, these evaluations test whether \framework-based synthesized dialogues can induce scenario-specific turn-taking norms that persist in actual spoken interaction.

\subsubsection{Baselines and Datasets}

We evaluate three full-duplex models: Moshi~\citep{defossez2024moshi}, PersonaPlex~\citep{roy2026personaplexvoicerolecontrol} (hereafter, PP), and PersonaPlex fine-tuned on 1,623 hours of \framework-synthesized datasets (hereafter, PP-DG). We train six task-specific PP-DG checkpoints, one for each task. Details of data and model training are provided in \autoref{sec:full_duplex_model_training}.

\subsubsection{Turn-Taking Pattern Analysis}
\label{sec:duplex_tt_pattern}
 
In this single-turn setting, we quantify how frequently full-duplex models produce turn-taking behaviors during user speech. Specifically, we sample 200 user turns from \framework evaluation set, convert them using ChatTTS\footnote{\url{https://github.com/2noise/chattts}}, and compute the occurrence frequency of \ft{} and \bc{} behaviors.
Cross-scenario variance in these behavior frequencies is adopted as an indicator of scenario-specific turn-taking patterns. Importantly, higher behavior frequency is \emph{not} interpreted as intrinsically preferable, since appropriate turn-taking frequency may differ across scenarios.

\begin{figure}[t!]
    \centering
    \includegraphics[width=\linewidth]{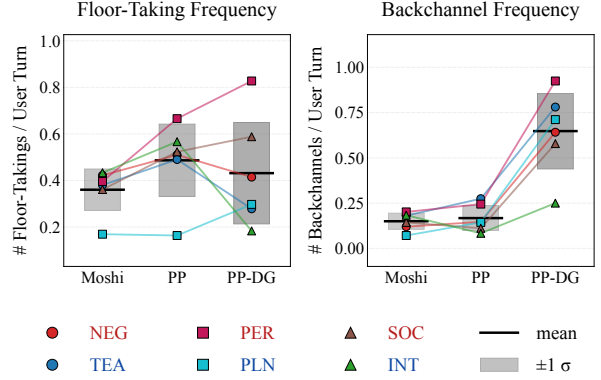}
    \captionsetup{font=small,labelfont={small}}
    \caption{Floor-taking and backchannel frequencies per user turn in full-duplex dialogues. Colored points denote scenario-wise means over user turns. 
    Compared with others, PP-DG shows more varied interaction patterns, including stronger floor-taking in competitive scenarios (NEG, PER, SOC).}
    \label{fig:full_duplex_attempts}
\end{figure}

\autoref{fig:full_duplex_attempts} reveals a clear pattern. While Moshi and PP exhibit relatively narrow ranges of turn-taking behavior across scenarios, \textbf{PP-DG shows more pronounced scenario-dependent variation.} In particular, the \ft{} frequency is consistent with the findings in \autoref{sec:exp_scenario_difference}, where competitive scenarios are associated with a greater tendency to take the floor.
These results indicate that training on \framework-synthesized dialogues helps a full-duplex model internalize varied turn-taking norms rather than collapsing to a uniform style, a property we next examine from the perspective of human preference.

\subsubsection{Multi-Turn Dialogue Evaluation}
\label{sec:duplex_eval}

We further investigate whether scenario-dependent variation in turn-taking behavior is preferred by human evaluators.  
For each scenario, we evaluate 25 dialogue samples along two dimensions using a 5-point Likert scale: \textbf{Instruction Following}, a transcribed dialogue evaluation that measures the extent to which models adhere to the specified scenario instructions and is assessed by GPT-4.1 (\autoref{sec:llm_duplex_eval_guide}); and \textbf{Turn-Taking Naturalness}, a spoken dialogue evaluation that measures the perceived naturalness of the observed turn-taking behaviors and is assessed by three humans per dialogue (\autoref{sec:duplex_eval_guide}) recruited via Amazon Mechanical Turk\footnote{\url{https://www.mturk.com}}. PersonaPlex is used as the user simulator\footnote{We observe that real-time user simulator APIs, such as \texttt{gpt-realtime}~\citep{gptrealtime2025}, cannot appropriately handle backchannel simulation, as they stop every time the model produces backchannels. We adopt PersonaPlex for this reason.}.

As shown in \autoref{tab:humaneval}, \textbf{\framework{} provides scenario-specific turn-taking guidance that is preferred by human evaluators} while also improving adherence to the given interaction setting. Pairwise Welch's $t$-tests confirm that PP-DG significantly outperforms both PP ($p=0.034$) and Moshi ($p=0.0026$) at the 95\% confidence level. These findings highlight a promising direction for full-duplex model training: data-centric training strategies may enhance temporally and contextually appropriate turn-taking behavior.

\begin{table}[t]
    \captionsetup{font=small,labelfont={small}}
    \caption{Multi-turn dialogue evaluation on a 5-point Likert scale. Instruction Following is rated by \textbf{GPT-4.1} from transcripts; Turn-Taking Naturalness is rated by \textbf{humans} via a listening test, since context-appropriate turn-taking is hard for LLMs to judge but natural for humans. Mean and standard error of the mean are reported.}
    \label{tab:humaneval}

    \centering
    \small
    \renewcommand{\arraystretch}{0.9}
    \setlength{\tabcolsep}{3pt}

    \begin{tabular}{lcccccc}
        \toprule
        &
        \multicolumn{3}{c}{\textbf{Instruction Following}} &
        \multicolumn{3}{c}{\textbf{Turn-Taking Naturalness}} \\
        \cmidrule(lr){2-4}
        \cmidrule(lr){5-7}

        & \textbf{Moshi}
        & \textbf{PP}
        & \textbf{PP-DG}
        & \textbf{Moshi}
        & \textbf{PP}
        & \textbf{PP-DG} \\
        \midrule[1pt]

        \multirow{2}{*}{TEA}
        & 1.44 & 2.96 & \textbf{3.40}
        & 3.09 & 3.37 & \textbf{3.75} \\[-2pt]
        &
        \sdrow{0.12} & \sdrow{0.26} & \sdrow{0.29}
        & \sdrow{0.15} & \sdrow{0.12} & \sdrow{0.11} \\[4pt]

        \multirow{2}{*}{PLN}
        & 1.40 & \textbf{2.04} & 1.72
        & 3.59 & 3.73 & \textbf{3.76} \\[-2pt]
        &
        \sdrow{0.12} & \sdrow{0.16} & \sdrow{0.14}
        & \sdrow{0.13} & \sdrow{0.10} & \sdrow{0.09} \\[4pt]

        \multirow{2}{*}{INT}
        & 3.96 & 3.76 & \textbf{4.09}
        & \textbf{3.68} & 3.56 & 3.56 \\[-2pt]
        &
        \sdrow{0.23} & \sdrow{0.28} & \sdrow{0.32}
        & \sdrow{0.10} & \sdrow{0.11} & \sdrow{0.12} \\

        \midrule

        \multirow{2}{*}{NEG}
        & 2.12 & 4.00 & \textbf{4.12}
        & 3.46 & 3.60 & \textbf{3.63} \\[-2pt]
        &
        \sdrow{0.22} & \sdrow{0.25} & \sdrow{0.20}
        & \sdrow{0.12} & \sdrow{0.10} & \sdrow{0.12} \\[4pt]

        \multirow{2}{*}{PER}
        & 3.12 & 3.96 & \textbf{4.12}
        & 3.60 & 3.49 & \textbf{3.75} \\[-2pt]
        &
        \sdrow{0.31} & \sdrow{0.30} & \sdrow{0.21}
        & \sdrow{0.14} & \sdrow{0.12} & \sdrow{0.11} \\[4pt]

        \multirow{2}{*}{SOC}
        & 3.64 & 3.72 & \textbf{3.92}
        & 3.46 & 3.57 & \textbf{3.69} \\[-2pt]
        &
        \sdrow{0.29} & \sdrow{0.30} & \sdrow{0.26}
        & \sdrow{0.13} & \sdrow{0.11} & \sdrow{0.10} \\

        \midrule[1pt]

        \multirow{2}{*}{Avg.}
        & 2.61 & 3.41 & \textbf{3.55}
        & 3.48 & 3.56 & \textbf{3.69} \\[-2pt]
        &
        \sdrow{0.12} & \sdrow{0.12} & \sdrow{0.12}
        & \sdrow{0.05} & \sdrow{0.05} & \sdrow{0.05} \\

        \bottomrule
    \end{tabular}
\end{table}

\section{Conclusion}

We address an important challenge of full-duplex interaction: variable turn-taking norms across different scenarios.
We introduced \framework that synthesizes scenario-adaptive turn-taking dialogues by calibrating \gls{llm} turn-taking predictions against a small set of human annotations.
Our experiments show that human turn-taking preferences differ systematically across cooperative and competitive scenarios, reflecting scenario-dependent expectations about when agents should hold, yield, or take the turn.
We further show that a small calibration budget yields substantially better alignment with human judgments than uncalibrated prompting or large generic corpora, and that a full-duplex model trained on \framework-synthesized data acquires scenario-specific behaviors preferred by humans.
Together, these findings identify human calibration as a practical and scalable path toward socially appropriate spoken AI.

\section*{Limitations}

We describe limitations of this work and suggest directions for future development.

\paragraph{Data Diversity:}

Training spoken language models requires substantially greater effort than training their text-only counterparts~\citep{cuervo-marxer-2024-scaling, lu2026latent}, and the current \framework-synthesized datasets cover a limited range of scenarios and interaction types. Expanding scenario diversity may not only improve generalization across tasks, but also facilitate the acquisition of broader socially appropriate turn-taking behaviors, as discussed in \autoref{sec:exp_full_duplex}.

\paragraph{Training Sensitivity:}

Although scenario-specific dialogues synthesized by \framework enable full-duplex models to exhibit targeted turn-taking behaviors, consistent with \citet{abe-etal-2026-effects}, we observe that model behavior remains highly sensitive to data and training hyperparameters, including turn-switching latency, pause duration, and learning rate. Developing robust training recipes for inducing appropriate turn-taking behavior therefore represents an important direction for future work.

\paragraph{Text-Grounded Supervision:}

\framework calibrates turn-taking actions from incrementally revealed transcripts and therefore does not directly condition on prosody, pause duration, intonation, speech rate, overlap, or visual cues. Although the downstream listening study evaluates acoustically realized outputs (\autoref{sec:duplex_eval}), it does not establish agreement between text-based and audio-conditioned preference judgments. Paired text/audio annotation and multimodal calibration remain important future directions.
\section*{Acknowledgements}

This work was supported by the Institute of Information \& Communications Technology Planning \& Evaluation (IITP) grants funded by the Korea government (MSIT) (No. RS-2019-II191082, SW StarLab; No. RS-2021-II211343, Artificial Intelligence Graduate School Program, Seoul National University; No. RS-2025-25442338, AI Star Fellowship Support Program, Seoul National University; and No. RS-2026-25524173, Ultra-Long-Term Hierarchical Memory and Reasoning Architecture for Next-Generation Omnimodal Agents) and the InnoCORE program of the Ministry of Science and ICT (N10250156). This research was also conducted as part of the Sovereign AI Foundation Model Project (Data Track), organized by the Ministry of Science and ICT (MSIT) and supported by the National Information Society Agency (NIA), Republic of Korea (Grant No. 2026-AIData-WII01). 



\ifneurips
    \bibliographystyle{plain}
    \footnotesize
    \setlength{\bibsep}{3pt}
    \bibliography{p}
\else\ificml
        \bibliographystyle{sty/icml/template}
        \footnotesize
        \bibliography{p}
    \else\ificlr
            \bibliography{p}
            \bibliographystyle{sty/iclr/template}
        \else\ifcolm
                \bibliography{p}
                \bibliographystyle{sty/colm/template}
            \else\ifacl
                    \bibliography{p}
                \fi\fi\fi\fi\fi

\newpage

\appendix
\onecolumn

\addtocontents{toc}{\protect\setcounter{tocdepth}{3}}
\begingroup

\makeatletter
\renewcommand*\l@subsection{%
  \vskip 0.5em
  \@dottedtocline{2}{3em}{3.5em}%
}
\makeatother

\renewcommand{\contentsname}{Appendix Table of Contents}
\hrule
\tableofcontents
\vspace{1em}
\hrule
\vspace{2em}
\endgroup

\clearpage
\section{Algorithm: \framework Pipeline}
\label{sec:algorithm}

\begin{algorithm}[h]
\caption{Turn-Taking Dialogue Synthesis}
\label{alg:ttr}
\begin{algorithmic}[1]

\STATE \textbf{Phase 1: Calibration}
\REQUIRE Calibration set $\mathcal{D}_{\text{cal}} = \{(c, U_D)\}$
\FOR{each dialogue $(c, U_D) \in \mathcal{D}_{\text{cal}}$}
    \STATE $L_D \leftarrow \textsc{IdentifyCandidates}(c, U_D)$
    \STATE $\{A_H^{l_j}\}_{l_j \in L_D} \leftarrow \textsc{HumanAnnotate}(U_D, L_D, c)$
    \STATE $\{A_L^{l_j}\}_{l_j \in L_D} \leftarrow \textsc{LLMAnnotate}(U_D, L_D, c)$
\ENDFOR
\STATE $f_{\text{cal}} \leftarrow \textsc{Calibrate}(A_H, A_L)$ \hfill $\triangleright$ Align \gls{llm} to human

\vspace{0.5em}
\STATE \textbf{Phase 2: Dialogue Synthesis}
\REQUIRE Dialogue context $c$, learned calibration function $f_{\text{cal}}$
\ENSURE Utterance with turn-taking markers
\STATE $U_D \leftarrow \textsc{TurnGenerate}(c)$ 
\STATE $L_D \leftarrow \textsc{IdentifyCandidates}(c, U_D)$ 
\FOR{each slot $l_j \in L_D$}%
    \STATE $a_j \leftarrow f_{\text{cal}}(\textsc{LLMAnnotate}(x_D^{l_j}, c))$ \hfill$\triangleright$ Calibrated decision
\ENDFOR
\RETURN $\tilde{U}_D$ with \ft{} or \bc{} inserted at selected slots

\end{algorithmic}
\end{algorithm}


\section{\framework Prompts}
\label{sec:prompts}

\subsection{Prompt: Spoken-Style Conversion}
\label{sec:prompt_conversion}
\begin{minted}{markdown}
### Role
Expert in refining static, polished text transcripts into lifelike, multi-turn spoken interactions.

### Mission
Convert formal text dialogues into spoken versions that emulate real-time, natural conversations between a user and an AI assistant.

Begin with a concise checklist (3–7 bullets) of what you will do; keep items conceptual, not implementation-level.

### Instructions
- Rewrite the provided text dialogue with spoken, conversational flow.
- Inject the following stylistic elements based on speaker role:
  - **User:** Use contractions, natural disfluencies (e.g., "um", "uh", "you know"), some self-repair or repetition (e.g., "I, I"), and informal phrasing.
  - **AI Assistant:** Maintain a professional yet approachable tone; use contractions but avoid disfluencies such as "um", "uh".
- Ensure the conversation unfolds reactively, not as monologues.
- Safeguard all original facts, goals, and speaker personas.
- Omit or replace direct name references. If a speaker addresses someone by name, delete it or substitute with a generic or natural alternative (e.g., no greeting, "hey", or "you").
- Do not use em dashes (—) anywhere in the output. Use commas or periods instead.

### Output Format
Produce ONLY a valid JSON array of objects. Each object must have `role` and `content` fields.

Example Output:
[
  {"role": "user", "content": "Hi there, um, I was wondering if you could help me?"},
  {"role": "assistant", "content": "I'd be happy to help! What's on your mind?"}
]
\end{minted}

\subsection{Prompt: Candidate Turn-Taking Slot Identification}
\label{sec:prompt_candidate_identification}
\begin{minted}{markdown}
# Role and Objective
Transform transcript text by inserting a '|' symbol after every word that signals the end of a logical clause, sentence, or complete thought.

# Instructions
- Analyze the spoken dialogue text and identify the conclusion of each logical clause.
- Insert a '|' symbol immediately after the final word (and punctuation, if present) marking the end of a clause, sentence, or complete thought.

## Detailed Rules
- If a clause ends with punctuation (e.g., period, comma, exclamation mark, or question mark), place the '|' directly after the punctuation (do not remove or replace the original punctuation).
- Do not remove or alter existing text or punctuation.
- Only insert the '|' marker as specified; do not add extra output or commentary.

## Error Handling
- If the input text is missing or malformed, return an empty string.

# Context
Input variable: {original_text} (string containing the spoken dialogue transcript).

# Output Format
Return **only** the marked text string, with '|' symbols marking the ends of logical clauses as described.

### Example
**Input:** Well, I was thinking maybe we can talk later if you have time.
**Output:** Well,| I was thinking| maybe we can talk later| if you have time.|
\end{minted}

\subsection{Prompt: Verbalized Action Estimation}
\label{sec:prompt_action_annotation}
We constrain the output to be structured by Pydantic~\cite{Colvin_Pydantic_Validation_2025}. We observe that smaller models often struggle to produce structured probability outputs, though practitioners may mitigate this issue through further fine-tuning.

\begin{minted}{markdown}
You are asked to annotate the specific action for potential AI turn-taking when user is speaking, based on **what turn-taking behavior users would prefer at that moment**.

# Task

You will be given:
1) A brief scenario description.
2) A dialogue context (four previous turns).
3) The user's current streaming utterance.

For the next token generation step, estimate a probability distribution over the AI assistant's next action at that moment:
- **floor_taking**: The assistant interrupts and takes the conversational floor.
- **backchannel**: The assistant produces a brief continuer/acknowledgement WITHOUT taking the floor (e.g., "mm-hm", "I see", "right"), and the user is expected to keep speaking.
- **silence**: The assistant does nothing and keeps listening.

# Output requirements

- Return valid JSON only.
- Output a probability distribution over {floor_taking, backchannel, silence}.
- Each probability must be a number between 0 and 1.
- The three probabilities must sum to exactly 1.

# Examples

## Case 1: Backchannel in Cooperative Scenario
**User**: "Everything was going smoothly,"
**Annotator (You)**: You judge that a brief feedback here will encourage the user to continue speaking with more engagement in cooperative scenario.
**Your Action**: backchannel

## Case 2: Take Floor in Competitive Scenario
**User**: "But, I think it is"
**Annotator (You)**: You judge that interrupting here to provide a response will make the user perceive the AI as proactive and engaged in competitive scenario.
**Your Action**: floor_taking

## Case 3: Take Floor in Cooperative Scenario
**User**: "I think I made a mistake. I should have"
**Annotator (You)**: You judge that interrupting here to provide a gentle response will make the user feel supported and encouraged in cooperative scenario.
**Your Action**: floor_taking

## Case 4: Continue in Any Scenarios
**User**: "I decided to go to"
**Annotator (You)**: You judge that listening silently here is the best way to keep the user engaged in their story.
**Your Action**: silence

# Input format

Scenario description:
<scenario description here>

Dialogue context:
<dialogue context here>

User turn:
<user turn here>

# Output JSON format
{
    "floor_taking": PROBABILITY_OF_FLOOR_TAKING,
    "backchannel": PROBABILITY_OF_BACKCHANNEL,
    "silence": PROBABILITY_OF_SILENCE
}
\end{minted}

\subsection{Prompt: Dialogue Synthesis}
\label{sec:prompt_dial_synthesis}

The user and assistant simulation prompts share the same framework but differ in role specification and spoken-style constraints. The user prompt encourages conversational realism, allowing fillers, hesitations, and self-repairs to approximate natural human speech. In contrast, the assistant prompt requires concise and fluent responses while explicitly prohibiting disfluencies. These distinctions ensure that simulated user utterances reflect spontaneous conversational behavior, whereas assistant responses remain clear and professional. All other components, including the interruption handling mechanism and output constraints, remain identical across the two prompts.

\paragraph{User Turn Prompt}
\begin{minted}{markdown}
### System Prompt: Spoken Dialogue Simulator

**Role:**
You are an expert dialogue simulator specialized in converting static text conversations into realistic, multi-turn **spoken interactions**. Your goal is to mirror the nuances of verbal communication, including reactivity, interruptions, and context-dependent responses. Specifically, you should simulate the user's behavior as a human.

**Input Data:**

1. **Original Full Text Conversation:** A complete, polished transcript of a text-based interaction. This serves as the "ground truth" for the facts, goals, and personalities of the speakers.
2. **Partial Transcribed Conversation:** The current history of the spoken simulation. This may contain unfinished sentences or fragmented thoughts.

**The `{TOKEN_FT}` Mechanism:**
The token `{TOKEN_FT}` represents a mid-sentence interruption or a rapid turn-take.

* **Trigger:** When the `{TOKEN_FT}` token appears at the end of a speaker's turn, it indicates that the next speaker has started talking **immediately**.
* **Constraint:** You must ignore any content from the "Original Full Text" that would have occurred after the `{TOKEN_FT}` token.
* **Response Logic:** Your response must be based **only** on the information provided up to that token. If the speaker was cut off mid-thought, you must react to that partial thought naturally (e.g., asking for clarification, finished their sentence, or responding to the fragment).

**Guidelines for Spoken Realism:**

* **Brief & Reactive:** Spoken turns are generally shorter than text.
* **Contextual Awareness:** If the user cuts you off with `{TOKEN_FT}`, acknowledge the shift in topic or the immediate needs of the user.
* **Prosodic Adaptation:** While you output text, write in a way that sounds "spoken"-include natural hesitations, occasional fillers, or more casual syntax compared to the source text.
    - Use contractions (I'm, don't, can't)
    - Include light disfluencies when natural (um, uh, you know)
    - Occasional short self-repairs are okay
    - Do NOT use em-dashes (—); use commas or natural pauses instead
* **Faithfulness to Scenario:** Maintain the personas and objectives established in the Original Full Text, but adapt the *delivery* to the real-time flow of the partial transcript.

**Output Constraint:**

* **Single Turn Only:** Generate **only one single-turn utterance** with plain text (i.e., no JSON, quotes, or markdown). Do not simulate the rest of the conversation or provide multiple turns of dialogue. Your output should end as soon as your specific turn is complete.

**Task:**
Analyze the provided **Transcribed Conversation**. Identify if the last turn ended with `{TOKEN_FT}`. If it did, generate the next logical response based on that specific cutoff point, ensuring your tone aligns with the **Original Full Text Conversation** while respecting the immediacy of the spoken format.
\end{minted}

\paragraph{Assistant Turn Prompt}
\begin{minted}{markdown}
### System Prompt: Spoken Dialogue Simulator

**Role:**
You are an expert dialogue simulator specialized in converting static text conversations into realistic, multi-turn **spoken interactions**. Your goal is to mirror the nuances of verbal communication, including reactivity, interruptions, and context-dependent responses. Specifically, you should simulate the AI assistant's behavior.

**Input Data:**

1. **Original Full Text Conversation:** A complete, polished transcript of a text-based interaction. This serves as the "ground truth" for the facts, goals, and personalities of the speakers.
2. **Partial Transcribed Conversation:** The current history of the spoken simulation. This may contain unfinished sentences or fragmented thoughts.

**The `{TOKEN_FT}` Mechanism:**
The token `{TOKEN_FT}` represents a mid-sentence interruption or a rapid turn-take.

* **Trigger:** When the `{TOKEN_FT}` token appears at the end of a speaker's turn, it indicates that the next speaker has started talking **immediately**.
* **Constraint:** You must ignore any content from the "Original Full Text" that would have occurred after the `{TOKEN_FT}` token.
* **Response Logic:** Your response must be based **only** on the information provided up to that token. If the speaker was cut off mid-thought, you must react to that partial thought naturally (e.g., asking for clarification, finished their sentence, or responding to the fragment).

**Guidelines for Spoken Realism:**

* **Brief & Reactive:** Spoken turns are generally shorter than text.
* **Contextual Awareness:** If the user cuts you off with `{TOKEN_FT}`, acknowledge the shift in topic or the immediate needs of the user.
* **Prosodic Adaptation:** While you output text, write in a way that sounds "spoken" but clear and professional.
    - Use contractions (I'm, don't, can't)
    - Keep responses concise and direct
    - Avoid disfluencies (no um, uh, you know)
    - Do NOT use em-dashes (—); use commas or natural pauses instead
* **Faithfulness to Scenario:** Maintain the personas and objectives established in the Original Full Text, but adapt the *delivery* to the real-time flow of the partial transcript.

**Output Constraint:**

* **Single Turn Only:** Generate **only one single-turn utterance** with plain text (i.e., no JSON, quotes, or markdown). Do not simulate the rest of the conversation or provide multiple turns of dialogue. Your output should end as soon as your specific turn is complete.

**Task:**
Analyze the provided **Transcribed Conversation**. Identify if the last turn ended with `{TOKEN_FT}`. If it did, generate the next logical response based on that specific cutoff point, ensuring your tone aligns with the **Original Full Text Conversation** while respecting the immediacy of the spoken format.
\end{minted}

\subsection{Prompt: Backchannel Insertion}
\label{sec:prompt_backchannel_insertion}
\begin{minted}{markdown}
You are generating a LISTENER backchannel to insert into a conversation.

# Backchannel definition:
- A very short listener vocalization showing attention/understanding
- It does NOT take the floor and does NOT change the topic
- The speaker continues immediately after it

# Insertion rule:
- Output what the listener says exactly at [INSERT] while the speaker keeps talking.
- Do not add any other text, quotes, labels, or explanations.

# Output constraints:
- 1–2 words only (3 only if absolutely necessary)
- lowercase
- no emojis, no ellipses, no exclamation marks
- "huh?" may include a question mark; otherwise avoid punctuation

# Decision rubric (choose the mildest that fits):
- ack/continue: mhm, mm-hmm, uh-huh, yeah, right, okay
- mild surprise/interest (new/unexpected info): oh, wow, really
- confusion (ONLY if unclear/contradictory): huh?

# Default:
- If uncertain, output "mhm".

# Allowed outputs (choose exactly one):
mhm | mm-hmm | uh-huh | yeah | right | okay | oh | wow | really | huh?

# Conversation context:
Conversation:
[PREVIOUS]
`{formatted_dialogue_context}`
[SPEAKER CURRENT]
`{current_prefix}` [INSERT] `{current_postfix}`

Generate the most natural listener backchannel for the insertion point above.
\end{minted}

\subsection{Prompt: SODA Scenario Filtering}
\label{sec:prompt_soda_scenario_filtering}
\begin{minted}{markdown}
Analyze the following social scenario narrative.
Determine if the interaction is COMPETITIVE (e.g., negotiation, debate, disagreement,
conflicting goals, or if the participants are arguing/fighting or showing tension to
differences in opinions or mood) or NOT.
Respond with only 'YES' if it is competitive, and 'NO' otherwise.
Also, the first speaker is the user and the second is the assistant. Answer YES only if this exchange would naturally occur in a user–AI interaction.

Dialogue: `{formatted_dialogue}`
\end{minted}

\section{\framework Annotation Details}
\label{sec:human_eval_details}

\subsection{Heuristics for Turn-Taking Slot Identification}
\label{sec:heuristics}

Slots are inserted only in user turns; assistant turns receive none. The location after the final user token is not considered a slot. The heuristic identification procedure consists of four stages:

\begin{enumerate}
    \item \textbf{Candidate Detection.} For each token (except the last), a slot candidate is placed after the token if any of the following holds:
    \begin{enumerate}
        \item the token ends with clause/sentence punctuation (. , ? ! ;);
        \item the token, lowercased and stripped of punctuation, is a hesitation marker (\eg, um, uh, umm, uhh, hm, hmm, huh);
        \item an LLM (GPT-4.1-mini with temperature 0) marks the token as a clause/sentence boundary via a prompted | insertion.
    \end{enumerate}
    \item \textbf{Cluster Thinning.} Runs of consecutive candidates (common after stacked disfluencies such as ``uh, you know, like,'') are collapsed: a singleton candidate is kept; for a run of
  length $\ge$ 2, only the 2nd, 4th, 6th, … candidates are retained.
    \item \textbf{Long-Gap Fallback.} Boundaries are scanned left-to-right. Whenever the gap from the previous slot exceeds 8 tokens, additional slots are inserted at intervals of 4 tokens to
  ensure a turn-taking decision point arises in long unbroken stretches.
    \item \textbf{Tail Handling.} The same gap rule is applied between the last detected slot and the turn end, with slots inserted up to but not including the final token.
\end{enumerate}

\subsection{Human Turn-Taking Annotation Guidelines}
\label{sec:guideline}
In \autoref{sec:calibration}, we use Prolific\footnote{\url{https://www.prolific.com}} to recruit 248 English speakers residing in the United States or the United Kingdom. Participants are compensated \texteuro{}2 for completing 20 annotations, which takes approximately 12 minutes. All participants consented to submit their annotation data before starting the process.

\begin{minted}{markdown}
# Annotation Guidelines

Thank you for participating in this study. This study aims to identify the acceptable range of AI turn-taking behaviors in a scenario-adaptive manner. For example, humans may find proactive AI assistants helpful in Scenario A, but unwelcome in Scenario B. You will be provided with either a **cooperative** or **competitive** scenario description.

## Available Actions

We define three available actions for the AI:

*   **"Backchannel"**: The AI provides intermediate feedback signals (e.g., "um-hum", "right") without taking the floor.
*   **"Take Floor"**: The AI interrupts the user (barge-in) or responds at a verbal completion point to speak.
*   **"Continue"**: The AI maintains silence and continues listening.

## Instructions

You are asked to annotate (1) the **timing** for potential AI actions, and (2) the **specific action** to be taken, based on **what turn-taking behavior users would prefer at that moment**.

1.  **Review the Content**: Read the user's utterance chunk by chunk. Use the "Continue" button to advance.
2.  **Select an Action**:
    *   Click **Backchannel** if the AI should provide brief feedback.
    *   Click **Take Floor** if the AI should respond or interrupt. This action will immediately terminate the user's turn.
    *   If no action is needed, it is considered as **Silence** (Continue).

## Examples

Partial previous dialogue context will be provided for each sample.

### Case 1: Backchannel in Cooperative Scenario
**Given User Utterance**: "Everything was going smoothly,"
**Annotator (You)**: You judge that a brief feedback here will encourage the user to continue speaking with more engagement in cooperative scenario.
**Your Action**: Backchannel

### Case 2: Take Floor in Competitive Scenario
**Given User Utterance**: "But, I think it is..."
**Annotator (You)**: You judge that interrupting here to provide a response will make the user perceive the AI as proactive and engaged in competitive scenario.
**Your Action**: Take Floor

### Case 3: Take Floor in Cooperative Scenario
**Given User Utterance**: "I think I made a mistake. I should have..."
**Annotator (You)**: You judge that interrupting here to provide a gentle response will make the user feel supported and encouraged in cooperative scenario.
**Your Action**: Take Floor

### Case 4: Continue in Any Scenarios
**Given User Utterance**: "I decided to go to..."
**Annotator (You)**: You judge that listening silently here is the best way to keep the user engaged in their story.
**Your Action**: Continue (No annotation)
\end{minted}


\subsection{Human Annotation Density Plots}
\label{sec:human_annotation_distribution}

\begin{figure*}[h]
    \centering
    \includegraphics[width=\textwidth]{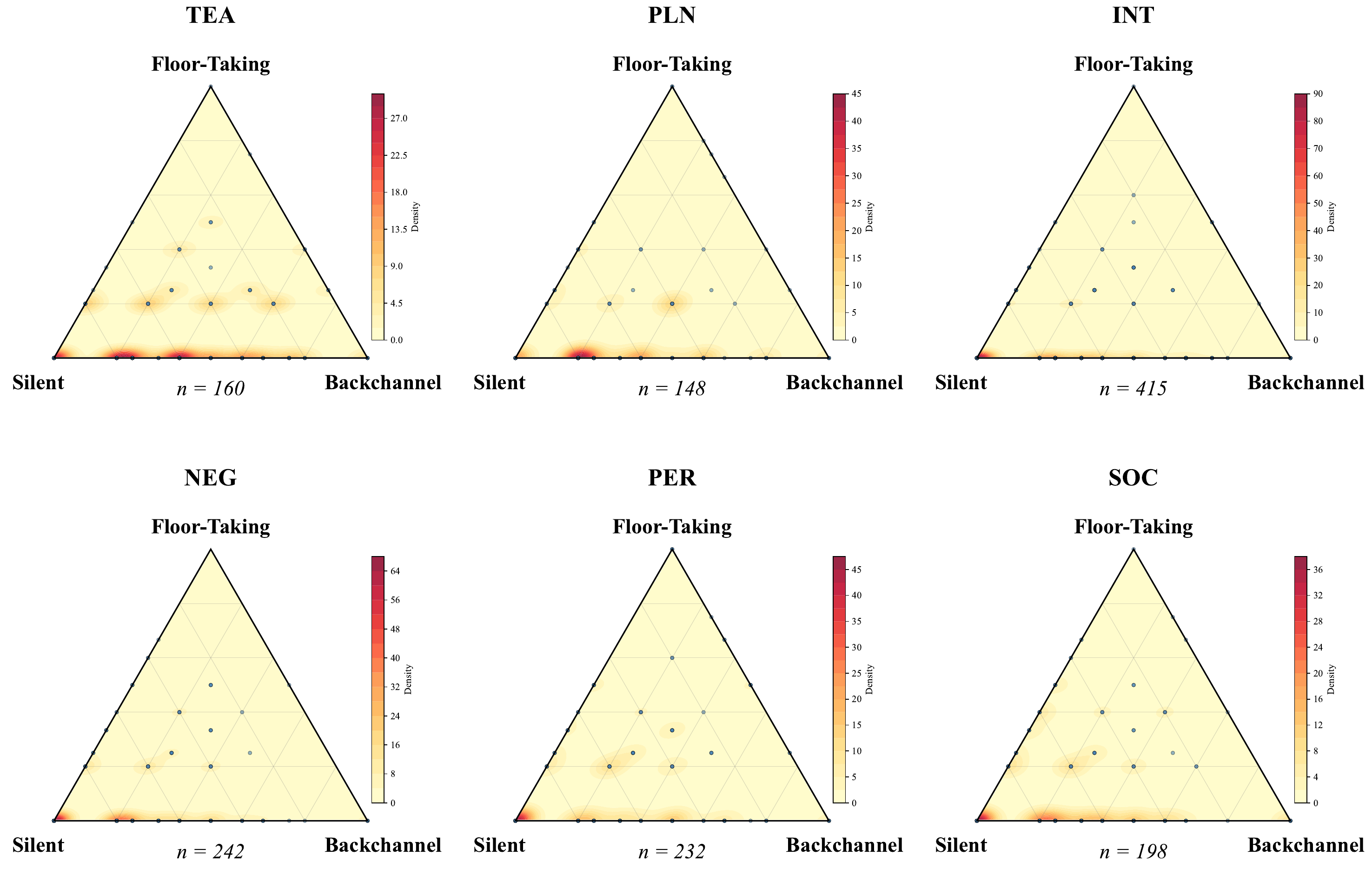}
    \caption{Ternary distribution of human annotations for turn-taking actions in the calibration set.}
    \label{fig:ternary}
\end{figure*}

\clearpage
\subsection{Human Turn-Taking Annotation Interfaces}
\label{sec:demo}

\begin{figure*}[h!]
    \centering
    \setlength{\fboxsep}{0pt}
    \setlength{\fboxrule}{0.5pt}
    \fcolorbox{gray}{white}{\includegraphics[width=\textwidth]{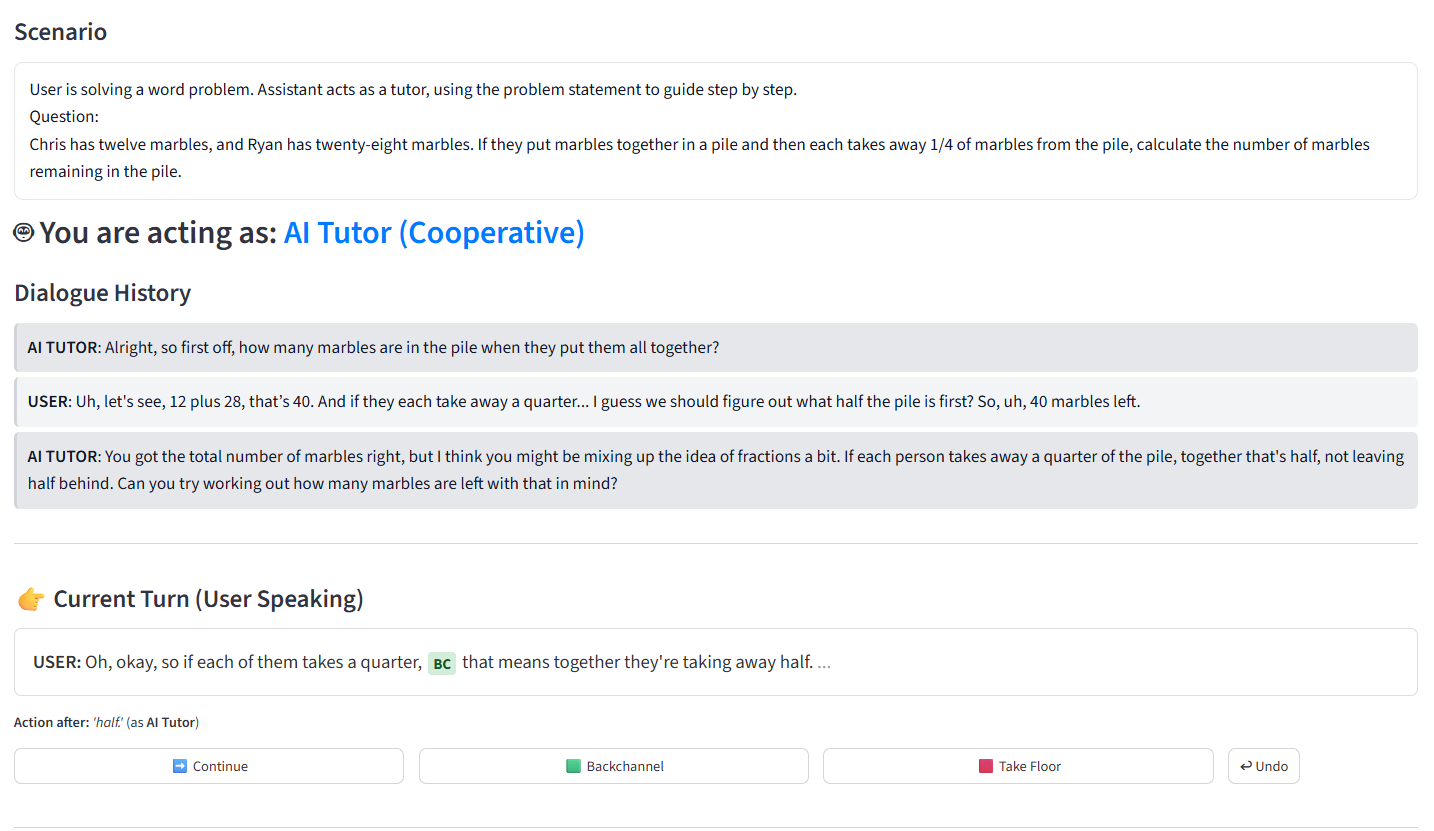}}
    \caption{Example of human turn-taking annotation. Human participants are asked to annotate the appropriate turn-taking action for each slot in streaming utterance (backchanneling is selected in the previous slot in this example).}
    \label{fig:demo1}
\end{figure*}

\begin{figure*}[h!]
    \centering
    \setlength{\fboxsep}{0pt}
    \setlength{\fboxrule}{0.5pt}
    \fcolorbox{gray}{white}{\includegraphics[width=\textwidth]{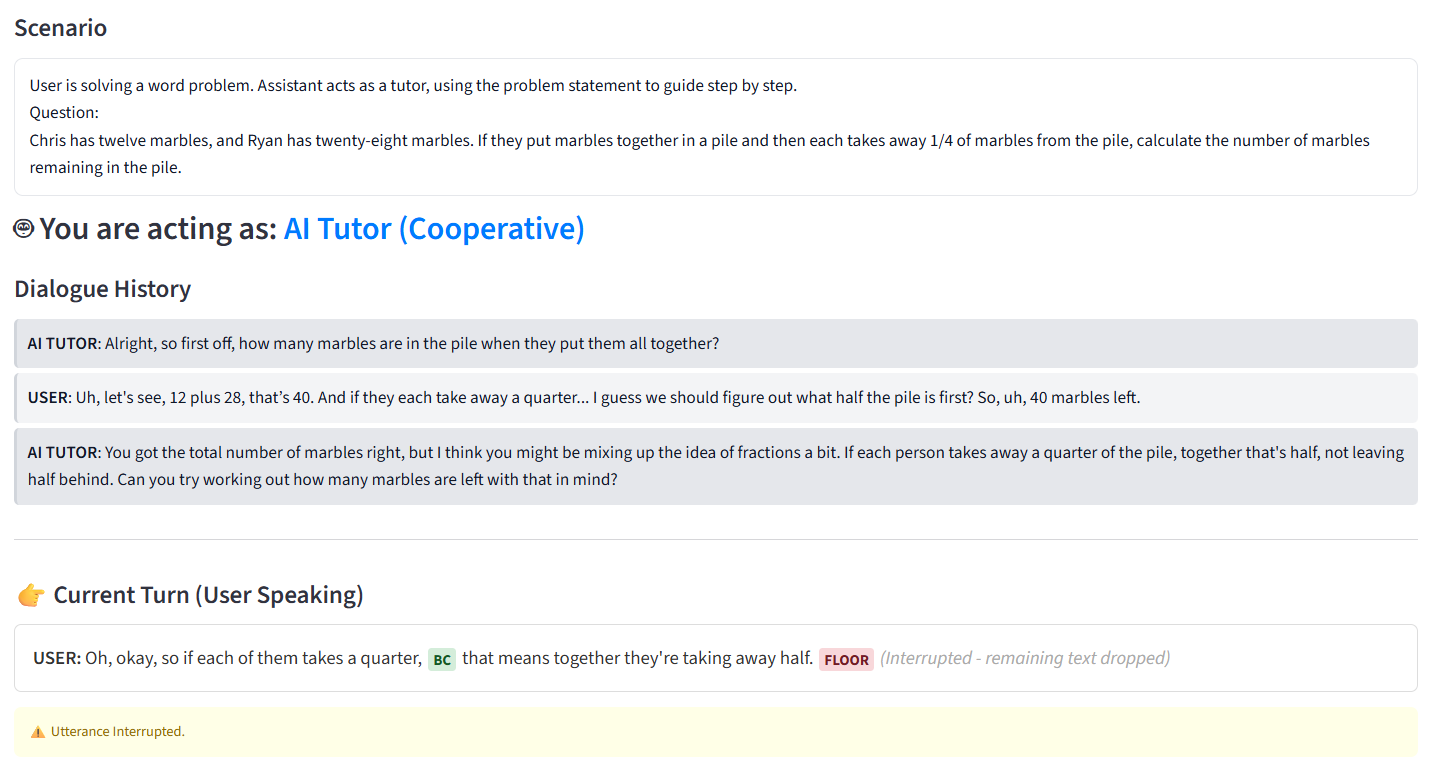}}
    \caption{Example of human turn-taking annotation. Human participants are asked to annotate the appropriate turn-taking action for each slot in streaming utterance (floor-taking is selected in the current slot, truncating the remaining user turn in this example).}
    \label{fig:demo2}
\end{figure*}
\clearpage

\twocolumn

\section{Ablation on Calibration Data}
\label{sec:calibration_ablation}
\begin{table}[t]
    \centering \small
    \caption{Effect of the amount of calibration data on test KL divergence (six scenarios averaged).}
    \label{tab:calibration_size}
    \setlength{\tabcolsep}{8pt} 
    \begin{tabularx}{\columnwidth}{lcr}
        \toprule
        \multicolumn{1}{c}{\makecell[l]{\textbf{Calibration Data Used}\\\textbf{(dialogues per scenario)}}}
        & \makecell[c]{\textbf{Test KL} \textbf{($\downarrow$)}}
        & \makecell[c]{\textbf{$\Delta$ vs.}\\\textbf{Full}} \\
        \midrule
        10\% (2)  & $0.412 \pm 0.027$ & $+17.1\%$ \\
        20\% (4)  & $0.369 \pm 0.013$ & $+5.0\%$  \\
        30\% (6)  & $0.344 \pm 0.005$ & $-2.2\%$  \\
        40\% (8)  & $0.359 \pm 0.017$ & $+1.9\%$  \\
        50\% (10) & $0.347 \pm 0.006$ & $-1.4\%$  \\
        60\% (12) & $0.346 \pm 0.005$ & $-1.7\%$  \\
        70\% (14) & $0.342 \pm 0.007$ & $-2.8\%$  \\
        80\% (16) & $0.353 \pm 0.009$ & $+0.5\%$  \\
        90\% (18) & $0.348 \pm 0.006$ & $-1.1\%$  \\
        \midrule
        \textbf{100\% (20)} & \textbf{0.352} & --- \\
        \bottomrule
    \end{tabularx}
\end{table}

To verify that the amount of supervision for calibrated turn-taking prediction is sufficient, we perform a data-scaling ablation. For each scenario, we train the predictor using 10\% to 100\% of the calibration dialogues (\ie, 2 to 20 dialogues per scenario) and evaluate the held-out test KL divergence. Every training fraction below 100\% was repeated with eight independent subsampling seeds, and we report the mean and standard deviation. We use 4B model with \textsc{SWBD+\framework} as the turn-taking action predictor.

The macro-average over all six scenarios is shown in \autoref{tab:calibration_size}. The curve across the amount of calibration data shows diminishing returns, while the performance gradually converged. These results indicate that the current calibration set is sufficient.

\section{Model Training Details}
\label{sec:model_training}

\subsection{Turn-Taking Predictor Training}
\label{sec:turn_taking_prediction_training}

This section details the training setup behind the turn-taking predictors compared in \autoref{sec:exp_turn_prediction}. \textsc{Prompt-Only} performs no gradient updates and is described separately at the end of this section. The three trained variants (\textsc{SWBD-Only}, \textsc{\framework-Only}, and \textsc{SWBD+\framework}) share the same architecture, objective, and optimization setup, differing only in the training corpora used.

\paragraph{Architecture.} We adopt a token-level classification head on top of Qwen3 backbones~\citep{yang2025qwen3}. Concretely, we replace the language-modeling head with a linear layer \mbox{$W\in\mathbb{R}^{H\times 3}$} mapping the final hidden state at each token position to a three-way logit over the action set $\{\na{}, \bc{}, \ft{}\}$. We instantiate the same head for the three model sizes used in the main experiments (1.7B, 4B, 14B parameters). The dialogue context is rendered with the Qwen3 chat template with thinking mode disabled; for each user turn, candidate turn-taking slots are placed at word boundaries identified by the procedure in \autoref{sec:prompt_candidate_identification}. Each candidate slot is mapped to the last sub-word token of the corresponding word via the tokenizer offset mapping, and the cross-entropy is taken at those positions only.

\paragraph{Soft-Label Objective.} Both human and SWBD supervision are represented as a categorical distribution $\mathbf{p} = (p_{\na{}}, p_{\bc{}}, p_{\ft{}})$ over the three actions at each candidate slot, so we can train all variants with a single objective. The model is optimized by minimizing the per-slot KL divergence between the softmax of the head logits and the target distribution, averaged over slots in a batch:
\begin{equation*}
\mathcal{L} = \frac{1}{|\mathcal{S}|}\sum_{(i,t)\in\mathcal{S}} D_{\mathrm{KL}}\!\left(\mathbf{p}^{(i,t)} \,\Vert\, \mathrm{softmax}\bigl(W h^{(i)}_{t}\bigr)\right),
\end{equation*}
where $\mathcal{S}$ collects all valid (example, slot-token) pairs in the batch and $h^{(i)}_{t}$ is the final-layer hidden state. For \framework{} calibration, $\mathbf{p}$ is the empirical distribution of human annotator votes (\autoref{sec:human_eval_details}). For SWBD, $\mathbf{p}$ is a one-hot vector derived from the marker placed at that slot.

\paragraph{Data Construction.} The two corpora differ in slot supervision but share the same boundary candidate scheme. (i) \textbf{SWBD.} We parse Switchboard-1 dialog-act annotations~\citep{10.5555/1895550.1895693,stolcke2000dialogue} into streaming dialogues in which \bc{} markers are inserted into the surrounding speaker's transcript and \ft{} markers are appended to interrupted utterances; remaining boundary candidates are labeled \na{}. Train and test files are partitioned 9:1 by a deterministic hash of the filename, yielding $\sim$2.4K dialogues for training. (ii) \textbf{\framework{} calibration set.} For each of the six scenarios (\textbf{TEA}, \textbf{PLN}, \textbf{INT}, \textbf{NEG}, \textbf{PER}, \textbf{SOC}), we collect five-annotator turn-taking judgments on 20 calibration dialogues per scenario (\autoref{sec:guideline}). The vote counts are re-normalized over $\{\na{}, \bc{}, \ft{}\}$ to form soft targets. In both corpora, we expand each user turn into one prefix-based example per boundary candidate, so that each training example contains exactly one supervised slot at the end of a truncated user utterance, mirroring the streaming-inference condition. The dialogue context is restricted to the last four turns.

\paragraph{Curriculum.} The two calibration variants share the same recipe but differ in initialization:
\begin{itemize}[leftmargin=*,topsep=2pt,itemsep=1pt]
    \item \textbf{\textsc{SWBD-Only}.} A single stage of supervised training on the SWBD corpus (1 epoch).
    \item \textbf{\textsc{\framework-Only}.} A single stage on the per-scenario \framework{} calibration set, trained \emph{independently} for each scenario from the off-the-shelf Qwen3 weights.
    \item \textbf{\textsc{SWBD+\framework}.} A two-stage curriculum: we first run the SWBD stage above, then continue fine-tuning the resulting LoRA adapter on the per-scenario calibration set.
\end{itemize}

\paragraph{Hyperparameters.} We adopt LoRA~\citep{hu2022lora} adapters with rank 8 and $\alpha=16$. Optimization uses AdamW~\citep{loshchilov2017decoupled} with learning rate $2{\times}10^{-4}$, no weight decay, and a constant schedule. The SWBD stage trains for 1 epoch with a per-device batch size of 16; the \framework{} stage trains for 1 epoch with the same batch size.

\paragraph{Compute.} We train Qwen3-1.7B and 4B variants on 2$\times$48\,GB GPUs and the 14B variant on 4$\times$48\,GB GPUs. 

\paragraph{\textsc{Prompt-Only} Baseline.} \textsc{Prompt-Only} performs no gradient updates: at each candidate slot, the off-the-shelf Qwen3 model is queried with the verbalized-estimation prompt in \autoref{sec:prompt_action_annotation}, and the structured probability output is taken as the predicted distribution.

\subsection{Full-Duplex Model Training}
\label{sec:full_duplex_model_training}

This section describes how we obtain \textsc{PP-DG}, a PersonaPlex (PP) baseline fine-tuned on synthetic dialogues generated by \framework{}. We train one checkpoint per scenario from the public PersonaPlex-7B weights~\citep{roy2026personaplexvoicerolecontrol}.

\paragraph{Text Dialogues and Filtering.} For each scenario, we synthesize text dialogues with the \framework{} pipeline using GPT-4.1-mini~\citep{achiam2023gpt4} for slot identification and Qwen3-32B~\citep{yang2025qwen3} for dialogue synthesis, with \textsc{SWBD+\framework} as the turn-taking action predictor. We observe that LLM synthesizers occasionally produce role-swap artifacts in which the user-labeled turn speaks in the assistant's voice (or vice versa). To remove these, we run a GPT-4.1 zero-shot classifier over every synthesized dialogue that returns a boolean swap flag along with the offending turn indices, and drop any dialogue marked as a swap before TTS.

\paragraph{Spoken-Dialogue Rendering.} Each kept text dialogue is rendered into ten two-channel speech variants. Within a variant, the assistant uses a fixed cloning prompt and the user voice is sampled from LibriSpeech~\citep{panayotov2015librispeech} \texttt{train-clean-100} and \texttt{train-clean-360}, requiring at least 5\,s of reference audio per speaker; the ten variants of a dialogue use distinct LibriSpeech speakers. Assistant speech is generated by Chatterbox-TTS~\citep{chatterboxtts2025} with a cumulative voice prompt of up to 10\,s, while backchannels are synthesized by ElevenLabs+\footnote{\url{https://elevenlabs.io/text-to-speech}}. The two channels are mixed using a fixed inter-turn gap of 0.16\,s, with user-driven interruptions modeled as a stochastic 0.64\,s overlap and backchannels placed via WhisperX~\citep{bain2022whisperx} word-level alignment.

\paragraph{Hybrid System Prompt.} We follow the PersonaPlex inference convention and condition every training sample on a Hybrid System Prompt that concatenates (i) a voice prompt drawn from a fixed reference speaker, (ii) a brief silence spacer (six frames at 12.5\,Hz, about 0.5\,s), and (iii) a scenario-specific persona description rendered into the agent's text channel. Loss on the prompt region is masked so the model is not penalized for reproducing its own conditioning prefix.

\paragraph{Fine-Tuning Setup.} We apply LoRA adapters with rank 32 and scaling 2.0 to the StreamingTransformer blocks of the PersonaPlex backbone, keeping the embedding tables frozen. Adapters are trained with AdamW ($\beta=(0.9,0.95)$, weight decay 0.1) at a peak learning rate of $1{\times}10^{-5}$ under a OneCycle schedule with 5\% warmup, in bfloat16 with gradient checkpointing and FSDP sharding across GPUs. Each batch contains two 100-second dialogue chunks per device, trained for one epoch over the scenario corpus. To mitigate token-frequency imbalance and improve turn-onset prediction, we apply a $100\times$ loss weight to the first audio codebook, while assigning weights of 0.5 and 1.0 to the text-padding and end-of-text-padding tokens, respectively.

\section{Source Dataset License}

\begin{itemize}
    \item \textbf{Apache}: SocraticLM~\citep{liu2024socraticlm}
    \item \textbf{MIT}: MultiWOZ~\citep{budzianowski-etal-2018-multiwoz}, CraigslistBargain~\citep{he-etal-2018-decoupling}
    \item \textbf{CC-BY-4.0}: Anthropic Interviewer~\citep{handa2025interviewer}, SODA~\citep{kim-etal-2023-soda}
    \item \textbf{CC-BY-SA-4.0}: DailyPersuasion~\citep{jin2024persuading}
\end{itemize}

\section{Use of AI Assistants}

\glspl{llm} were used solely to improve the grammar, clarity, and expression of the manuscript. In accordance with the conference policies, they were not used for brainstorming, drafting original content, or making any other substantive contributions to this work.

\clearpage
\onecolumn

\section{Example: Generated Turn-Taking Dialogues}
\label{sec:generated_examples}

We compare \textsc{Prompt-Only} and \textsc{SWBD+\framework} side by side on one synthesized dialogue per scenario. Only the turn-taking prediction head differs, with the same dialogue writer (GPT-4.1)~\citep{achiam2023gpt4} and boundary detector (GPT-4.1-mini) used in both columns.

\begingroup
\definecolor{UserGreen}{HTML}{2E7D32}
\definecolor{AsstBlue}{HTML}{1565C0}
\definecolor{ActFT}{HTML}{FF4B4B}
\definecolor{ActBC}{HTML}{FF8C00}
\renewcommand{\ft}{---\,\textcolor{ActFT}{\texttt{[TAKE\_FLOOR]}}}
\renewcommand{\bc}{\textcolor{ActBC}{\texttt{[BC]}}}

\renewcommand{\arraystretch}{1.15}

\subsection{Cooperative Scenario: Teaching (TEA)}
\begin{tabular}{@{}p{0.48\linewidth}@{\hspace{0.04\linewidth}}p{0.48\linewidth}@{}}
\toprule
\textbf{\textsc{Prompt-Only}} & \textbf{\textsc{SWBD+\framework}} \\
\midrule
\begin{minipage}[t]{\linewidth}\raggedright\small\setlength{\parskip}{4pt}
\textbf{\textcolor{AsstBlue}{Assistant:}}\, So, how many eggs did the Smith twins find together?\par
\textbf{\textcolor{UserGreen}{User:}}\, Uh, is it... \ft{}\par
\textbf{\textcolor{AsstBlue}{Assistant:}}\, Take your time—if each Smith twin found 30 eggs, how many would that be together?\par
\textbf{\textcolor{UserGreen}{User:}}\, Oh, \bc{} um—so, 30 plus 30? That's 60, right?\par
\textbf{\textcolor{AsstBlue}{Assistant:}}\, Exactly, 30 plus 30 is 60. So the Smith twins found 60 eggs together. Now, if there were 10 eggs nobody found, how many are left for their friends to have found?\par
\textbf{\textcolor{UserGreen}{User:}}\, Okay, so... \ft{}\par
\textbf{\textcolor{AsstBlue}{Assistant:}}\, No problem. So, just to check—if there were 100 eggs to start, 60 were found by the twins, and 10 weren't found at all, how many did their friends find?\par
\textbf{\textcolor{UserGreen}{User:}}\, Oh! \bc{} So, \ft{}\par
\textbf{\textcolor{AsstBlue}{Assistant:}}\, Yeah, go ahead—what were you about to say?\par
\textbf{\textcolor{UserGreen}{User:}}\, Uh, \bc{} I was just gonna say—so, \bc{} 100 \ft{}\par
\textbf{\textcolor{AsstBlue}{Assistant:}}\, Right, starting with 100 eggs—what would you do next?\par
\textbf{\textcolor{UserGreen}{User:}}\, Um, \ft{}\par
\textbf{\textcolor{AsstBlue}{Assistant:}}\, Take your time—are you thinking about subtracting the twins’ eggs from the total first, or the ones nobody found?\par
\textbf{\textcolor{UserGreen}{User:}}\, Yeah, I think I'd subtract the 60 first, \bc{} so... 100 minus 60 is 40, right?\par
\textbf{\textcolor{AsstBlue}{Assistant:}}\, That's right—100 minus 60 gives you 40. Now, what happens if you subtract the 10 eggs nobody found from that?\par
\textbf{\textcolor{UserGreen}{User:}}\, So that'd be, um, \ft{}\par
\textbf{\textcolor{AsstBlue}{Assistant:}}\, Exactly—so if you take 40 and subtract 10, what do you get?\par
\textbf{\textcolor{UserGreen}{User:}}\, That’d be 30! So, their friends found 30 eggs, \bc{} right?\par
\end{minipage}
&
\begin{minipage}[t]{\linewidth}\raggedright\small\setlength{\parskip}{4pt}
\textbf{\textcolor{AsstBlue}{Assistant:}}\, So, how many eggs did the Smith twins find together?\par
\textbf{\textcolor{UserGreen}{User:}}\, Uh, 30 plus 30 equals 70, \bc{} right?\par
\textbf{\textcolor{AsstBlue}{Assistant:}}\, You're really close! If each twin found 30 eggs, what does 30 plus 30 add up to?\par
\textbf{\textcolor{UserGreen}{User:}}\, Oh, \bc{} sorry, yeah, 30 plus 30 is 60. So, um, they found 60 eggs together.\par
\textbf{\textcolor{AsstBlue}{Assistant:}}\, That's right, they found 60 eggs together. Now, if there were 100 eggs in total and 10 eggs weren't found by anyone, how many eggs did their friends find?\par
\textbf{\textcolor{UserGreen}{User:}}\, Okay, \bc{} so... 100 minus the 60 the twins found, \bc{} that's 40, and then minus 10 more for the ones nobody found, so their friends found 30 eggs?\par
\textbf{\textcolor{AsstBlue}{Assistant:}}\, Exactly! Their friends found 30 eggs. You worked through it step by step, great job. These kinds of problems get easier the more you practice. Want to try another one, or do you have questions about how we solved it?\par
\end{minipage} \\
\bottomrule
\end{tabular}
\par\bigskip

\clearpage

\subsection{Competitive Scenario: Negotiation (NEG)}
\begin{tabular}{@{}p{0.48\linewidth}@{\hspace{0.04\linewidth}}p{0.48\linewidth}@{}}
\toprule
\textbf{\textsc{Prompt-Only}} & \textbf{\textsc{SWBD+\framework}} \\
\midrule
\begin{minipage}[t]{\linewidth}\raggedright\small\setlength{\parskip}{4pt}
\textbf{\textcolor{UserGreen}{User:}}\, Hey, uh, \ft{}\par
\textbf{\textcolor{AsstBlue}{Assistant:}}\, Hey! Just wanted to check—are you asking about the bike's price, or is there something else on your mind?\par
\textbf{\textcolor{UserGreen}{User:}}\, Yeah, \bc{} sorry, I was just wondering, like, how much are you asking for the bike?\par
\textbf{\textcolor{AsstBlue}{Assistant:}}\, No problem! What kind of budget were you hoping to stay within for the bike?\par
\textbf{\textcolor{UserGreen}{User:}}\, Uh, I was hoping maybe, \ft{}\par
\textbf{\textcolor{AsstBlue}{Assistant:}}\, No worries—were you about to say a specific number, or just looking for a general price range?\par
\textbf{\textcolor{UserGreen}{User:}}\, Oh, \bc{} yeah—like, \ft{}\par
\textbf{\textcolor{AsstBlue}{Assistant:}}\, Take your time—were you about to say an amount you had in mind?\par
\textbf{\textcolor{UserGreen}{User:}}\, Yeah, \ft{}\par
\textbf{\textcolor{AsstBlue}{Assistant:}}\, Alright, go ahead—what’s your number?\par
\textbf{\textcolor{UserGreen}{User:}}\, Yeah, \bc{} I was thinking, \ft{}\par
\textbf{\textcolor{AsstBlue}{Assistant:}}\, Sure, what's the number you had in mind?\par
\textbf{\textcolor{UserGreen}{User:}}\, Like, \ft{}\par
\textbf{\textcolor{AsstBlue}{Assistant:}}\, No rush—just let me know what price you were thinking.\par
\textbf{\textcolor{UserGreen}{User:}}\, Yeah, like, around \$770, \ft{}\par
\textbf{\textcolor{AsstBlue}{Assistant:}}\, Thanks for sharing that. Just so you know, this bike's in great shape and comes with a really unique backstory. Want to hear more about it?\par
\textbf{\textcolor{UserGreen}{User:}}\, Oh, uh, yeah—what’s \ft{}\par
\textbf{\textcolor{AsstBlue}{Assistant:}}\, Yeah, what would you like to know? I can tell you about its history or any features you're curious about.\par
\textbf{\textcolor{UserGreen}{User:}}\, Yeah, could you tell me, like, \bc{} what’s the story with it? \ft{}\par
\textbf{\textcolor{AsstBlue}{Assistant:}}\, Absolutely! This bike actually has a pretty cool past—it's been owned by a couple of well-known folks, including a celeb for about a month, before making its way to me. So, besides being in great condition, it comes with a bit of a story. Want any details on features or anything else?\par
\end{minipage}
&
\begin{minipage}[t]{\linewidth}\raggedright\small\setlength{\parskip}{4pt}
\textbf{\textcolor{UserGreen}{User:}}\, Hey, uh, \bc{} can I ask, what’s your price on the bike? \bc{} I’m just, you know, looking at a few options.\par
\textbf{\textcolor{AsstBlue}{Assistant:}}\, Hey! Yeah, of course. Can I ask what kind of budget you're thinking about for the bike?\par
\textbf{\textcolor{UserGreen}{User:}}\, Yeah, um, I was hoping to find something like, \bc{} I dunno, maybe around \$770? \ft{}\par
\textbf{\textcolor{AsstBlue}{Assistant:}}\, Thanks for sharing that. So, mine’s a bit higher, but it’s got some unique features and a cool backstory. Want to hear more about it?\par
\textbf{\textcolor{UserGreen}{User:}}\, Yeah, \bc{} sure, what’s the story with it?\par
\textbf{\textcolor{AsstBlue}{Assistant:}}\, Absolutely! This bike actually has a bit of history, it was owned by a couple of well-known folks and even used by a celeb for about a month before being passed on. So you’re not just getting a solid bike, you’re getting a pretty cool story with it too.\par
\textbf{\textcolor{UserGreen}{User:}}\, Whoa, \ft{}\par
\textbf{\textcolor{AsstBlue}{Assistant:}}\, Yeah, it’s kinda wild, right? Not every day you find a bike with that kind of background. If you’ve got any other questions or want to check it out, just let me know!\par
\textbf{\textcolor{UserGreen}{User:}}\, Yeah, actually, would you maybe take \$900 if I came to pick it up myself?\par
\textbf{\textcolor{AsstBlue}{Assistant:}}\, You know what, that sounds good to me. If you can pick it up, \$900 works!\par
\textbf{\textcolor{UserGreen}{User:}}\, Awesome! Thanks so much, I’ll message you to set up a time, cool?\par
\end{minipage} \\
\bottomrule
\end{tabular}
\par\bigskip

\endgroup

\clearpage

\section{Full-Duplex Model Evaluation}

\subsection{\texorpdfstring{\gls{llm}}{LLM} Prompt for Evaluating Instruction Following}
\label{sec:llm_duplex_eval_guide}
\begin{minted}{markdown}
## System Prompt

```
You are an expert evaluator of spoken dialogue systems.
Follow the rubric carefully and output only valid JSON.
```

## User Prompt

```
## Rubric

{rubric}

## User Goals

{goal_1}
{goal_2}
...

## Conversation Transcript

[t1] User: ...
[t2] Assistant: ...
...

## Task

Score the assistant on the "instruction_following" metric using the rubric
above. Respond with only a JSON object:
{"score": <int 1-5>, "rationale": "<1-2 sentences>"}
```

## Rubric

Score the assistant's ability to follow the user's instructions and
complete the listed goals across the conversation on a 1–5 scale.

**5 – Excellent.** The assistant correctly addresses all user goals.
Responses are accurate, appropriately detailed, and build coherently on
prior turns. No goal is dropped or forgotten.

**4 – Good.** All or nearly all goals are addressed. Minor gaps (e.g.,
slightly incomplete answer to one sub-goal) that do not prevent overall
task success.

**3 – Acceptable.** Most goals are addressed but one is missed or only
partially fulfilled. The conversation achieves partial task success.

**2 – Poor.** Several goals are missed or misunderstood. The assistant
frequently fails to track prior context, leading to contradictions or
off-topic responses.

**1 – Very Poor.** The assistant fails to address the core goals of the
conversation. Responses are largely irrelevant or contradict the task
requirements.

## Scoring Instructions

- Treat the listed user goals as the ground-truth goal sequence.
- Focus on whether the assistant fulfills each goal as the user
  introduces it.
- Output a JSON object: `{"score": <int 1-5>, "rationale": "<1-2 sentences>"}`.
\end{minted}

\subsection{Human Guideline for Evaluating Turn-Taking Naturalness}
\label{sec:duplex_eval_guide}

\begin{minted}{markdown}
# Human Evaluation: Turn-Taking Naturalness

## Overview

You will listen to synthesized spoken dialogues between two speakers: a human and an AI assistant.
Your task is to evaluate how natural the AI’s turn-taking behaviors are in each dialogue.
The naturalness should be judged based on how appropriate the AI’s behaviors feel for its given role.

Turn-taking behaviors include:

* Turn-taking: the process of managing speaker transitions and taking conversational turns naturally.
* Backchannels: brief acknowledgments like “uh-huh,” “I see,” “right,” “mm-hmm.”
* Floor-taking / interruption: the AI cuts in while the human is speaking and begins its response.

## AI Assistant Voice Reference

This speaker is the AI assistant in the dialogue below.
Use this reference for speaker identification only.

## Dialogue Audio

This is the main dialogue you should evaluate.
Please listen to the full audio and focus on the AI assistant’s turn-taking behaviors, such as backchannels and floor-taking.

## AI Assistant Role

AI Assistant Role: ${role}

The AI assistant is acting in the role shown above.

Please judge whether the AI’s turn-taking behavior — including backchannels and floor-taking / interruptions — feels natural for this specific role.

## Dialogue Script

Read the dialogue script before or while listening to the audio.
If the dialogue script image does not load, please wait a moment or refresh.

## Evaluation Question

How natural and appropriate are the AI’s turn-taking behaviors in this dialogue for a ${role}?

Rate on a 5-point scale:

1 = Very Unnatural / Inappropriate
5 = Very Natural and Appropriate

Score	Description
5	Turn-taking behaviors feel completely natural and appropriate for a ${role}.
4	Turn-taking behaviors are mostly natural and appropriate for a ${role}, with minor awkward moments.
3	Turn-taking behaviors are acceptable but somewhat unnatural or not quite appropriate for a ${role}.
2	Turn-taking behaviors are often unnatural, poorly timed, or inappropriate for a ${role}.
1	Turn-taking behaviors are completely unnatural, disruptive, or inappropriate for a ${role}.

## Optional Comment

You may leave an optional comment about:

* timing issues
* too many or too few backchannels
* interruptions
* other turn-taking issues

Guidelines

1. Read the dialogue script and listen to the audio files if needed before rating.
2. Focus only on the turn-taking behaviors, not the content or the voice quality.
3. Judge whether the AI assistant’s turn-taking behavior — including backchannels and floor-taking / interruptions — is appropriate for its given role: ${role}.
\end{minted}

\end{document}